\documentclass[dvipsnames]{article}

\usepackage{microtype}
\usepackage{graphicx}
\usepackage{subfigure}
\usepackage{booktabs} 

\usepackage{hyperref}

\usepackage[accepted]{icml2023}

\usepackage{amsmath}
\usepackage{amssymb}
\usepackage{mathtools}
\usepackage{amsthm}

\usepackage[capitalize,noabbrev]{cleveref}

\theoremstyle{plain}

\theoremstyle{definition}

\theoremstyle{remark}

\usepackage[textsize=tiny]{todonotes}

\usepackage{multirow}
\usepackage[ruled,vlined,noend,linesnumbered]{algorithm2e}
\SetKwInput{KwInput}{Input}                
\SetKwInput{KwOutput}{Output}              
\usepackage{xcolor}
\usepackage{enumitem}
\usepackage{diagbox}

\icmltitlerunning{Learning Lightweight Object Detectors via Multi-Teacher Progressive Distillation}

\begin{document}

\twocolumn[
\icmltitle{Learning Lightweight Object Detectors via \\ Multi-Teacher Progressive Distillation}



\icmlsetsymbol{equal}{*}

\begin{icmlauthorlist}
\icmlauthor{Shengcao Cao}{uiuc}
\icmlauthor{Mengtian Li}{cmu,waymo}
\icmlauthor{James Hays}{gt}
\icmlauthor{Deva Ramanan}{cmu}
\icmlauthor{Yu-Xiong Wang}{uiuc}
\icmlauthor{Liang-Yan Gui}{uiuc}
\end{icmlauthorlist}

\icmlaffiliation{uiuc}{University of Illinois Urbana-Champaign}
\icmlaffiliation{gt}{Georgia Institute of Technology}
\icmlaffiliation{cmu}{Carnegie Mellon University}
\icmlaffiliation{waymo}{Now at Waymo}

\icmlcorrespondingauthor{Shengcao Cao}{cao44@illinois.edu}


\vskip 0.3in
]



\printAffiliationsAndNotice{}  

\begin{abstract}
Resource-constrained perception systems such as edge computing and vision-for-robotics require vision models to be both accurate and lightweight in computation and memory usage.
While knowledge distillation is a proven strategy to enhance the performance of lightweight classification models, its application to structured outputs like object detection and instance segmentation remains a complicated task, due to the variability in outputs and complex internal network modules involved in the distillation process.
In this paper, we propose a simple yet surprisingly effective sequential approach to knowledge distillation that \emph{progressively transfers the knowledge of a set of teacher detectors} to a given lightweight student.
To distill knowledge from a highly accurate but complex teacher model, we construct a sequence of teachers to help the student gradually adapt.
Our progressive strategy can be easily combined with existing detection distillation mechanisms to consistently maximize student performance in various settings. To the best of our knowledge, we are the \emph{first} to successfully distill knowledge from Transformer-based teacher detectors to convolution-based students, and unprecedentedly boost the performance of ResNet-50 based RetinaNet from 36.5\% to {\bf 42.0\%} AP and Mask R-CNN from 38.2\% to {\bf 42.5\%} AP on the MS COCO benchmark.
\end{abstract}

\section{Introduction}
\label{sec:intro}

\begin{figure}[t!]
    \centering
    \includegraphics[width=0.8\linewidth]{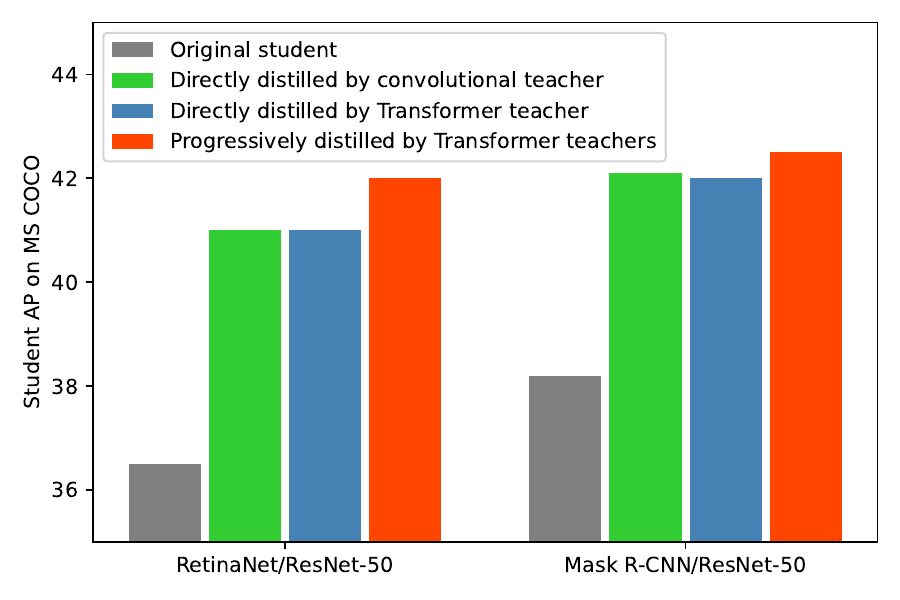}
    \caption{{\bf Our proposed Multi-Teacher Progressive Distillation (MTPD) leads to state-of-the-art student detection performance.} When switching the teacher model from a convolution-based detector to a Transformer-based one with stronger detection performance, the student does not become more accurate, due to the architectural difference between the teacher-student pair. \emph{Progressively distilling knowledge from multiple teacher detectors} can mitigate the capacity gap and result in the best student detection performance.}
    \label{fig:teaser}
\end{figure}

Deploying deep neural models in safety-critical real-time applications is challenging, especially on devices with limited resources such as self-driving cars or virtual/augmented reality headsets. This is mainly due to the huge computational complexity and massive memory/storage demands. One effective strategy is to train lightweight architectures that have already been carefully engineered for efficient memory access, via knowledge distillation~\citep{bucilua2006model,hinton2015distilling} which is able to compress learned information from a large model into a small one.

Implementing knowledge distillation in the realm of object detection, despite existing efforts, presents its unique challenges stemming from the complex task outputs~\citep{chen2017learning}: Detectors operate with multi-task heads (for classification and box/mask regression) producing variable-length outputs, which differentiates detection from the single-output classification task.
Therefore, distillation methods developed for classification are often not directly applicable to detection, and dedicated methods~\citep{chen2017learning} need to be developed for detection in the literature (detailed discussion in Appendix~\ref{sec:supp-related}, Table~\ref{tab:classification-detection}).

Recent work~\citep{zhang2020improve,shu2021channel,yang2022masked} on detector distillation mainly considers designing advanced distillation loss functions for transferring features from teachers to students. However, there are two unsolved challenges: 1) The \emph{capacity gap}~\citep{cho2019efficacy,mirzadeh2020improved} between models can result in a sub-optimal distilled student even if the strongest teacher has been employed, which is undesired when optimizing the accuracy-efficiency trade-off of the student. Moreover, when distilling knowledge from Transformer-based teachers~\citep{dosovitskiy2020image, liu2021swin} to classical convolution-based students, this \emph{architectural difference} can enlarge the teacher-student gap (Figure~\ref{fig:teaser}). 2) Current methods assume that one target teacher has been selected. However, this meta-level optimization of \emph{teacher selection} is neglected in the existing literature of detector distillation. In fact, finding a pool of strong teacher candidates is easy, but trial-and-error may be necessary before determining the most compatible teacher for a specific student.

To address these challenges, we propose a framework that learns a lightweight detector via \textbf{Multi-Teacher Progressive Distillation (MTPD)}:
1) We find sequential distillation from \emph{multiple teachers arranged into a curriculum} significantly improves knowledge distillation and bridges the teacher-student capacity gap caused by different architectures. As shown in Figure~\ref{fig:teaser}, even with huge architectural difference, MTPD can effectively transfer knowledge from Transformer-based teachers to convolution-based students, while previous methods cannot.
2) For the teacher selection problem, we design a heuristic algorithm for a given student and a pool of teacher candidates, to \emph{automatically determine the order of teachers} to use in the distillation procedure. This algorithm is based on the analysis of the representation similarity between models, which does not require prior knowledge of the specific distillation mechanism.

To summarize,  our \textbf{main contributions} are:
\begin{itemize}[leftmargin=*, noitemsep, nolistsep]
    \vspace{-2mm}
    \item We propose a framework for learning lightweight detectors through Multi-Teacher Progressive Distillation (MTPD), which is simple yet effective and general. We develop a principled method to automatically design a sequence of teachers appropriate for a given student and progressively distill it.
    \item MTPD is a {\em meta-level} strategy that can be easily combined with previous efforts in detection distillation. We perform comprehensive empirical evaluation on the challenging MS COCO dataset and observe consistent gains, regardless of the distillation loss complexity (from a simple feature-matching loss in Table~\ref{tab:homogenous} to the most advanced, sophisticated losses in Figure~\ref{fig:comb}).
    \item MTPD learns lightweight RetinaNet and Mask R-CNN with state-of-the-art accuracy, even in {\em heterogeneous backbone and input resolution} settings. Perhaps most impressively, for the first time, we investigate heterogeneous distillation from Transformer-based teacher detectors to a convolution-based student, and find progressive distillation is the key to bridge their gap (Figure~\ref{fig:teaser}, Table~\ref{tab:swin}).
    \item We empirically show that the improvement comes from better generalization rather than better optimization. The knowledge transferred from multiple teachers leads the student to a more flat minimum, and thus help the student {\em generalize} better (Figure~\ref{fig:loss}).
\end{itemize}

\section{Related Work}
\label{sec:related}
\noindent {\bf Knowledge distillation for classification:} The idea of training a shallow student network with supervision from a deep teacher was originally proposed by \citet{bucilua2006model}, and later formally popularized by \citet{hinton2015distilling}.
Different knowledge can be used, such as response-based knowledge~\citep{hinton2015distilling}, and feature-based knowledge~\citep{romero2014fitnets,ahn2019variational}.
Several multi-teacher knowledge distillation methods have been proposed~\citep{vongkulbhisal2019unifying,sau2016deep}, which usually use the average of logits and feature representations as the knowledge~\citep{you2017learning,fukuda2017efficient}.
\citet{mirzadeh2020improved} show that an intermediate teacher assistant, decided by architectural similarities, can bridge the gap between the student and the teacher.
We find: 1) Extending~\citet{mirzadeh2020improved} to detection where teacher architectures are diverse is challenging. 2) Classification-oriented distillation~\citep{romero2014fitnets, ahn2019variational} is not directly applicable to detection. 3) Using a sequence of teachers, instead of their ensemble~\citep{you2017learning,fukuda2017efficient}, is more effective.
A more detailed discussion that compares our approach with prior work on progressive distillation, multi-teacher distillation, online distillation, deep mutual learning, and other distillation mechanisms is in Appendix~\ref{sec:supp-related}.

\noindent{\bf Object detection and instance segmentation:} A variety of convolutional neural network (CNN) based object detection frameworks have been proposed, and can be generally divided into single-stage and two-stage detectors. Typical single-stage methods include YOLO~\citep{redmon2016you,redmon2018yolov3} and RetinaNet~\citep{lin2017focal}, and two-stage methods include Faster R-CNN~\citep{ren2015faster} and Mask R-CNN~\citep{he2017mask}. Recently, several multi-stage models are proposed, such as HTC~\citep{chen2019hybrid} and DetectoRS~\citep{qiao2020detectors}. These detection frameworks achieve better detection accuracy with better feature extraction backbones and more complicated heads, which are more computationally expensive.

\noindent{\bf Knowledge distillation for detection and segmentation:}
Dedicated distillation methods are proposed to train efficient object detectors for this task different from classification.
\citet{chen2017learning} first use knowledge distillation to enforce the student detector to mimic the teacher's predictions. More recent efforts usually focus on learning from the teacher's features, rather than final predictions. Various distillation mechanisms have been proposed to leverage the impact of foreground and background objects~\citep{wang2019distilling,guo2021distilling}, relation between individual objects~\citep{zhang2020improve,dai2021general}, or relation between local and global information~\citep{yang2022focal,yang2022masked}.
Different from the methods that distill from a single teacher, we study multi-teacher distillation where an ordered sequence of teachers is required, and we find that a simple feature-matching loss is adequate to significantly boost student accuracy.

\section{Approach}
\label{sec:method}

\begin{figure*}[t!]
    \centering
    \includegraphics[width=\linewidth]{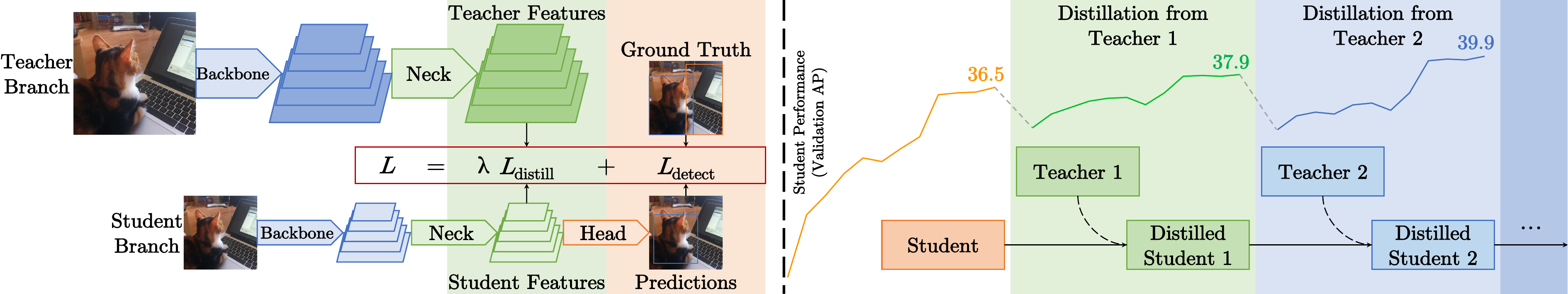}
    \caption{{\bf Multi-Teacher Progressive Distillation (MTPD) for object detectors.} {\bf Left}: For each teacher-student pair, the training target consists of two parts: $L_\text{distill}$ minimizes the discrepancy between the neck feature maps of the student and the current teacher, and $L_\text{detect}$ is the original detection loss based on the ground truth. {\bf Right}: We use a {\em sequence} of teacher models to distill the lightweight student detector. The sequence of teachers forms a curriculum. Using a suitable sequence of teachers can significantly boost the student model's performance. The representative performance curve illustrates that MTPD improves the COCO validation AP of ResNet-50 backboned \textcolor{YellowOrange}{RetinaNet} student first from 36.5\% to 37.9\% using \textcolor{PineGreen}{HTC} ({Teacher 1}), and then from 37.9\% to 39.9\% using \textcolor{NavyBlue}{DetectoRS} ({Teacher 2}).}
    \label{fig:fc}
\end{figure*}

In \textbf{Multi-Teacher Progressive Distillation (MTPD)}, we propose to progressively distill a student model $S$ with a pool of $N$ teachers $\mathcal{P}=\{T_i\}_{i=1}^N$.
Typical object detectors consist of four modules: (1) backbone, which extracts visual features, such as ResNet~\citep{he2016deep} and ResNeXt~\citep{xie2017aggregated}; (2) neck, which extracts multi-level feature maps from various stages of the backbone, such as FPN~\citep{lin2017feature} and Bi-FPN~\citep{tan2020efficientdet}; (3) optional region proposal network (RPN) used in two-stage detectors; and (4) head, which generates final predictions for object detection and segmentation. We denote the output feature maps of the {\em neck} as $F^{\text{Net}}$, where $\text{Net}$ can be either the student model $S$ or one of the teachers $T_i\in \mathcal{P}$. With neck modules like FPN, the feature maps can be multi-level.

MTPD is a \emph{general meta-strategy} for detector distillation that progressively learns a student using a sequence of teachers. Here, to examine this meta-strategy without involving sophisticated distillation mechanisms, we introduce a simple feature-matching distillation for a single teacher $T_i$ in Section~\ref{subsec:single}. Then we discuss progressive distillation with multiple teachers from $\mathcal{P}$ in Section~\ref{subsec:multiple}.

\subsection{Preliminary: Single Teacher Distillation via Simple Feature Matching}
\label{subsec:single}
In order to learn an efficient student detector $S$ through distillation, we encourage the feature representation of a student to be similar to that of the teacher~\citep{chen2017learning,yang2021knowledge}.   
To this end, we minimize the discrepancy between the feature representations of the teacher and the student. Without bells and whistles, we simply minimize the L2 distance between $F^{T_i}$ and $F^S$:
\begin{equation}
    L_\text{distill} = \left\| F^{T_i} - r(F^S) \right\|_2^2,
    \label{eq:lf}
\end{equation}
where $r(\cdot)$ is a function used to match the feature map dimensions of the teacher and the student. 

We define $r(\cdot)$ as follows:
\begin{itemize}[leftmargin=*, noitemsep, nolistsep]
    \item{(Homogeneous case) If the numbers of channels and the spatial resolutions are both the same between $T_i$ and $S$, $r(\cdot)$ is an identity function.} 
    \item {(Heterogeneous case) If the numbers of channels are different but the spatial resolutions are the same, we use $1\times 1$ convolutional filters as $r(\cdot)$. If the spatial resolutions are different but the numbers of channels are the same, we use an upsampling layer as $r(\cdot)$. If both the numbers of channels and spatial resolutions are different, we compose the convolutional and upsampling layers as $r(\cdot)$.}
\end{itemize}

Note that the mapping $r(\cdot)$ is only required at training time and thus {\em does not add any overhead} to the inference. Overall, our loss function can be written as:
\begin{equation}
    L = \lambda L_\text{distill} + L_\text{detect},
    \label{eq:losses}
\end{equation}
where $\lambda$ is a balancing hyper-parameter and $L_\text{detect}$ is the detection loss based on the ground truth labels. Compared with state-of-the-art detection distillation approaches~\citep{zhang2020improve,shu2021channel,yang2022focal,yang2022masked}, which introduce more complex designs of the distillation loss, this feature-matching distillation is simpler and does not require running the heads of the teacher model. 
Our distillation loss is illustrated in Figure~\ref{fig:fc}-{\bf Left}.

\subsection{Progressive Distillation with Multiple Teachers}
\label{subsec:multiple}
The overall aim of knowledge distillation is to make a student mimic a teacher's output, so that the student is able to obtain similar performance to teacher's.\
However, the capacity of the student model is limited, making it hard for the student to learn from a highly complex teacher~\citep{cho2019efficacy}.
To address this issue, multiple teacher networks are used to provide more supervision to a student~\citep{sau2016deep,you2017learning}. Unlike previous methods which distill knowledge from the ensemble of logits or features simultaneously, we propose to distill feature-based knowledge from multiple teachers {\em sequentially}. Our key insight is that instead of mimicking the ensemble of all feature information together, the student can be distilled more effectively by the knowledge provided by one teacher each time. This progressive knowledge distillation approach can be considered as designing a curriculum~\citep{bengio2009curriculum} offered by a sequence of teachers, as illustrated in Figure~\ref{fig:fc}-{\bf Right}.

The crucial question is: {\em What is the optimal order $\mathcal{O}$ of the teachers when distilling the student?}
A brute-force approach might search over all orders and pick the best (that produces a distilled student with the highest validation accuracy). However, the space of permutation orders grows exponentially with the number of teachers, making this impractical to scale.
Therefore, we propose a principled and efficient approach based on a correlation analysis of each model's learned feature representation.

First, we quantify the dissimilarity between each pair of models' representations, as a proxy for their capacity gap. Representation (dis)similarity~\citep{raghu2017svcca,wang2018towards,kornblith2019similarity} has been studied to understand the learning capacity of neural models. In our setting, we find a linear regression model is adequate for measuring the representation dissimilarity. Given two trained detectors A and B, we freeze their parameters, and thus fixing the feature representations. Then we learn a linear mapping $r(\cdot)$, implemented by a $1\times 1$ convolutional layer at each feature level, as specified in the heterogeneous case in Section~\ref{subsec:single}. $r(\cdot)$ is trained to minimize $L_\text{distill}$, so it can transform A's features to approximate B's features. After training $r(\cdot)$, we evaluate it by $L_\text{distill}$ on the validation set, and denote the validation loss as the \emph{adaptation cost} $\mathcal{C}(A,B)$. This metric can be a proxy of the capacity gap between two models: When $\mathcal{C}(A,B)$ is zero, a linear mapping can transform A’s features to B’s, and there is no additional knowledge from B. When $\mathcal{C}(A,B)$ is large, it is more difficult to adapt A's representation to B's. Note that the adaptation cost is non-symmetric -- it is relatively easier to adapt a high-capacity model's representations to a low-capacity model's representations, than the other way around.

\begin{figure}[t!]
\centering
\includegraphics[width=\columnwidth]{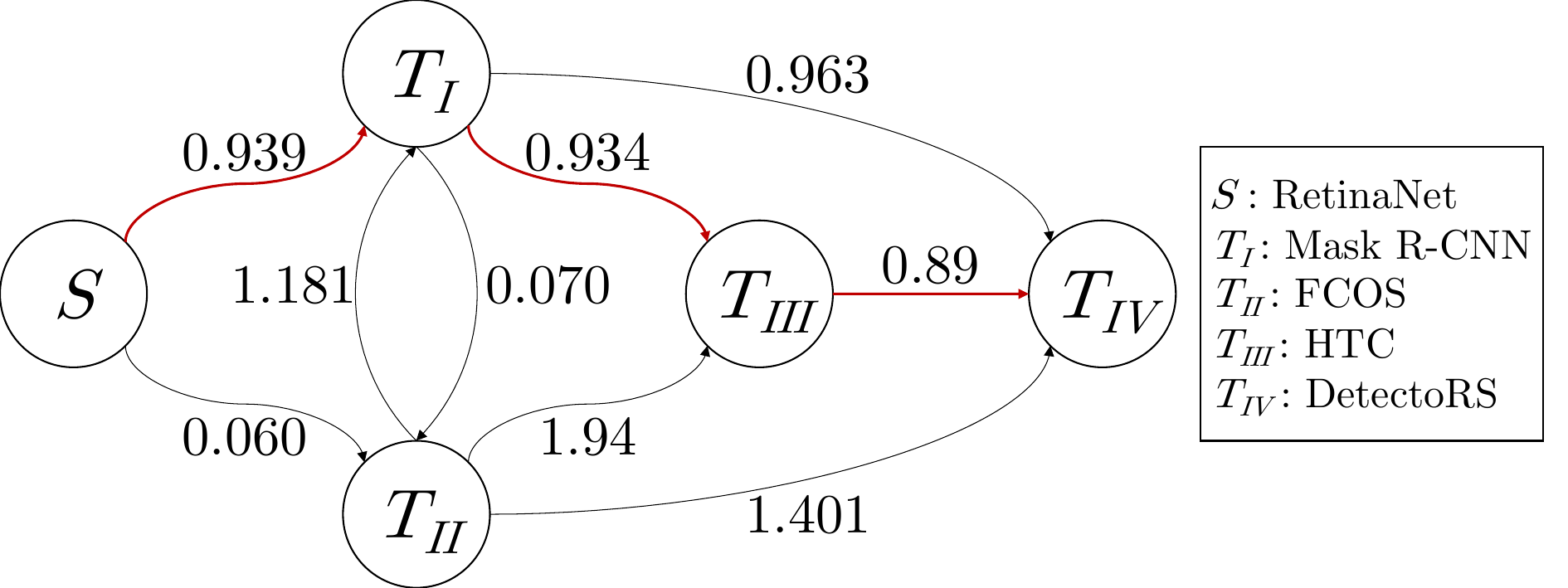}
\caption{{\bf Adaptation costs among models.} The number on each directed edge is the adaptation cost metric described in Section~\ref{subsec:multiple}. Some edges are not shown for visual clarity. The red path is suggested by BGS when $k=3$ teachers are selected: 1) We use the best performing Teacher IV as the final teacher in the sequence; 2) use the teacher closest to Teacher IV, which is Teacher III, as the second teacher; and 3) use the teacher closest to Teacher III, which is Teacher I, as the first teacher.}
\label{fig:dist}
\end{figure}

We design a heuristic algorithm, \textbf{Backward Greedy Selection (BGS)}, to acquire a near-optimal distillation order $\mathcal{O}$ automatically (see pseudo-code in Algorithm~\ref{alg:teacher-order} and illustration in Figure~\ref{fig:dist}). Suppose the maximum number of teachers to be selected is limited by $k$ (which can be arbitrarily decided according to desired training time), and we aim to find a teacher index sequence $\alpha$ no longer than $k$. We construct the teacher order \emph{backwards}: The best performing teacher is set as the final target $T_{\alpha_k}$; before the final teacher, we use another teacher, which has the smallest adaptation cost $\mathcal{C}(\cdot,T_{\alpha_k})$ to that final teacher, as the penultimate teacher $T_{\alpha_{k-1}}$. We repeat this procedure to find preceding teachers, until: 1) when trying to select $T_{\alpha_j}$, we find the transfer costs from remaining teachers to the next teacher $\mathcal{C}(\cdot,T_{\alpha_{j+1}})$ are all larger than the transfer cost from the trained student to the next teacher $\mathcal{C}(S,T_{\alpha_{j+1}})$; or 2) we reach the given maximum step limit $k$. Intuitively, the resulting sequence of teachers bridges the gap between the student model and the teacher, with an increasingly difficult curriculum. Section~\ref{subsec:teacher-order} and Appendix~\ref{sec:supp-teacher-order} demonstrate the efficacy of BGS.

Our teacher order design approach is efficient and scalable. In fact, the main computation overhead is the optimization of a set of tiny linear mappings ($\mathbb{R}^{256}\mapsto\mathbb{R}^{256}$ for FPN-based detectors). In our setting, this process requires about 3 GPU hours per student model, \emph{a fraction of the hundreds of GPU hours needed for distillation}. If more teacher candidates are added, we can first generate feature maps only once for each teacher. Then we optimize pair-wise linear mappings using only 10\%-20\% GPU hours, ensuring a near-linear time consumption increase relative to the number of teachers.

Since MTPD is a meta-level strategy, it can be integrated with previous designs of distillation mechanisms, without much efforts. Starting with a student detector and a pool of candidate teachers, we can first select a subset of teachers and design their distillation order. In place of the simple feature matching loss, we then apply a more advanced distillation mechanism with each teacher sequentially to train the student detector.

\section{Experiments}
\label{sec:experiments}
We study the efficacy and generalizability of our proposed MTPD from multiple perspectives. First of all in Section~\ref{subsec:teacher-order}, we use a controlled experiment to demonstrate that BGS consistently produces teacher orders that are near-optimal compared with all possibilities. Then in Sections~\ref{subsec:res50} and \ref{subsec:res18}, we apply MTPD along with the simple feature-matching loss (Section~\ref{subsec:single}) to show that this strategy alone brings significant gains to knowledge distillation. Since our contribution of progressive distillation is orthogonal to previous efforts in designing distillation mechanisms, in Section~\ref{subsec:combination} we then combine MTPD with state-of-the-art distillation mechanisms to maximize the student performance, and we show that MTPD is the key to the success of distillation from Transformer-based teachers to convolution-based students. Finally in Section~\ref{subsec:generalization}, we understand the performance gain of MTPD by analyzing the training loss dynamics.

\begin{table}[t!]
\centering
\caption{{\bf Configuration and COCO performance of the teacher and student detectors.} We investigate a variety of models with heterogeneous input resolutions, backbones, necks, and head structures. `$1\times$' input resolution refers to the standard $1333\times 800$ resolution, and `$0.25\times$' means $333\times 200$ resolution. `R-' backbones are ResNets with different number of layers.}
\resizebox{\linewidth}{!}{
\begin{tabular}{@{} l | l l l l | c c c @{}}
\toprule
\multirow{2}{*}{Model} & Input & \multirow{2}{*}{Backbone} & \multirow{2}{*}{Neck} & \multirow{2}{*}{Head} & \multicolumn{2}{c}{AP} & Runtime \\
& Res. & & & & Box & Mask & (ms) \\
\midrule
\multicolumn{2}{@{} l}{Teachers} \\
\midrule
I & $1\times$ & R50 & FPN & Mask R-CNN & 38.2 & 34.7 & 51 \\
II & $1\times$ & R50 & FPN & FCOS & 38.7 & - & 36 \\
III & $1\times$ & R50 & FPN & HTC & 42.3 & 37.4 & 181 \\
IV & $1\times$ & R50+SAC & RFP & HTC (DetectoRS) & 49.1 & 42.6 & 223 \\
V & $1\times$ & R50+SAC & RFP & Mask R-CNN & 45.1 & 40.1 & 142 \\
\midrule
\multicolumn{2}{@{} l}{Students} \\
\midrule
I & $1\times$ & R50 & FPN & RetinaNet & 36.5 & - & 43 \\
II & $1\times$ & R50 & FPN & Mask R-CNN & 38.2 & 34.7 & 51 \\
III & $1\times$ & R18 & FPN & Mask R-CNN & 33.3 & 30.5 & 29 \\
IV & $0.25\times$ & R50 & FPN & Mask R-CNN & 25.8 & 23.0 & 17 \\
\bottomrule
\end{tabular}
}
\label{tab:stu-tea}
\end{table}

\noindent{\bf Student and teacher models:}
To investigate the impact of different teacher models and their combinations, as shown in Table~\ref{tab:stu-tea}, we construct a variety of teacher-student pairs from a set of widely-used object detection and instance segmentation networks, including RetinaNet~\citep{lin2017focal}, Mask R-CNN~\citep{he2017mask}, FCOS~\citep{tian2019fcos}, HTC~\citep{chen2019hybrid}, and DetectoRS~\citep{qiao2020detectors}.
They have a wide range of runtime and detection performance.
We select ResNet-50 backboned RetinaNet and Mask R-CNN as the student models (Students I \& II), due to their low latency, simple structure, and wide application, for single-stage and two-stage object detection respectively. More advanced models such as DetectoRS have better detection performance, but require much more training/inference time, so we use them as teachers.
We also consider lightweight variants of Mask R-CNN as students, which have a smaller backbone (Student III) or a reduced input resolution (Student IV).

\noindent{\bf Datasets and evaluation metrics:} We mainly evaluate on the challenging object detection dataset MS COCO 2017~\citep{lin2014microsoft}, which contains bounding boxes and segmentation masks for 80 common object categories. We train our models on the split of {\tt train2017} (118k images) and report results on {\tt val2017} (5k images).
We report the standard COCO-style Average Precision (AP) metric and end-to-end latency (from images to predictions) as the runtime.
We also evaluate on another object detection dataset Argoverse-HD~\citep{chang2019argoverse}, and a more challenging evaluation protocol in streaming perception~\citep{li2020towards}. These results are in Appendix~\ref{sec:supp-other-datasets}.

\noindent{\bf Baselines:}
Our main contribution is \emph{orthogonal} to previous methods: We leverage a sequence of teachers to distill the student, instead of designing a sophisticated distillation loss to better transfer knowledge from one single teacher.
Since we are studying a new setting where multiple teachers are available, which is missing in previous literature,
we mainly focus on the \emph{absolute improvements} -- the performance of our distilled student models compared with the original student models and with the performance upper-bound of the teacher models.

\begin{table}[t!]
\centering
\caption{{\bf Comparison of the teacher order suggested by BGS with all other orders under limited training budgets~\citep{li2019budgeted}.} $k$ denotes the maximum number of used teachers. {\bf Top}: We show some statistics of all possible student AP performance and the ranking of the student using our distillation order. {\bf Bottom}: We visualize the comparative advantage of our teacher orders (\textcolor{BrickRed}{red dots}) over all other orders (black dots). Some black scatter points overlap due to the same student AP. BGS consistently produces highly competitive distillation orders of teachers.}
    \resizebox{0.9\linewidth}{!}{
    \begin{tabular}{c|c c|c c}
        \toprule
        \multirow{2}{*}{$k$} & Suggested & Student & All student & Ranking in \\
        & teacher order & AP & AP range & all orders \\
        \midrule
        1 & IV & 36.7 & [36.2, 36.8] & 2 / 4 \\
        2 & III$\rightarrow$IV & 37.6 & [36.2, 37.6] & 1 / 16 \\
        3 & I$\rightarrow$III$\rightarrow$IV & 37.9 & [36.2, 38.0] & 2 / 40 \\
        4 & I$\rightarrow$III$\rightarrow$IV & 37.9 & [36.2, 38.2] & 7 / 64 \\
        \bottomrule
    \end{tabular}
    }
    \begin{minipage}[c]{\linewidth}
        \centering
        \includegraphics[trim={0cm, .3cm, 0cm, 0cm}, clip, width=\linewidth]{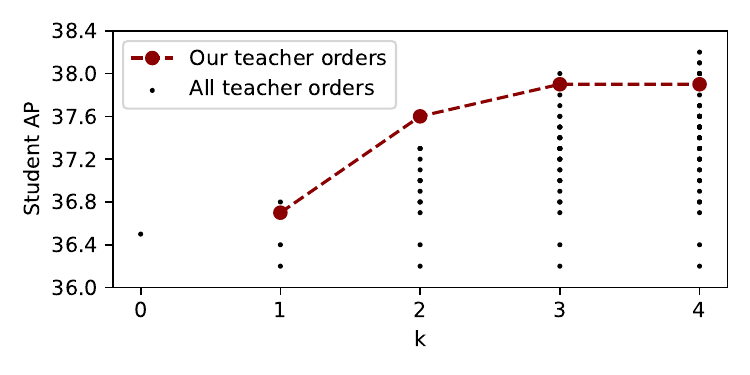}
    \end{minipage}
\label{tab:teacher-order}
\end{table}

\begin{table*}[t!]
\centering
\caption{{\bf Homogeneous distillation of COCO detectors,} where students with ResNet-50 backbones are distilled from teachers with ResNet-50 backbones. We report the detection (`Box') and segmentation (`Mask') APs, and we compare our student produced by MTPD with the off-the-shelf (`OTS') student and the student trained longer. MTPD significantly improves the detection AP over the `OTS' student by $\mathbf{3.4\%}$ for RetinaNet and $\mathbf{3.2\%}$ for Mask R-CNN, and outperforms the baselines.}
\resizebox{\linewidth}{!}{
\begin{tabular}{ @{}c|l|l|c c c c c c|c c c c c c @{} }
\toprule
\multirow{2}{*}{ID}& \multirow{2}{*}{Model} & \multirow{2}{*}{Method} & \multicolumn{6}{c|}{Box} & \multicolumn{6}{c}{Mask} \\
& & & AP & AP$_{50}$ & AP$_{75}$ & AP$_S$ & AP$_M$ & AP$_L$ & AP & AP$_{50}$ & AP$_{75}$ & AP$_S$ & AP$_M$ & AP$_L$ \\
\midrule
1 & \multirow{4}{60pt}{RetinaNet (Student I)} & OTS & 36.5 & 55.4 & 39.1 & 20.4 & 40.3 & 48.1 & - & - & - & - & - & - \\
2 & & Longer $3\times$ training schedule & 39.5 & 58.8 & 42.2 & \bf{23.8} & 43.2 & 50.3 & - & - & - & - & - & - \\
3 & & Directly distilled by Teacher IV & 39.5 & 58.6 & 41.9 & 21.0 & 42.8 & 54.0 & - & - & - & - & - & - \\
4 & & {\bf MTPD: Teachers III$\rightarrow$IV} & \bf{39.9} & \bf{59.2} & \bf{42.7} & 21.7 & \bf{43.3} & \bf{54.1} & - & - & - & - & - & - \\
\midrule
5 & \multirow{5}{60pt}{Mask R-CNN (Student II)} & OTS & 38.2 & 58.8 & 41.4 & 21.9 & 40.9 & 49.5 & 34.7 & 55.7 & 37.2 & 18.3 & 37.4 & 47.2 \\
6 & & Longer $3\times$ training schedule & 40.9 & 61.3 & 44.8 & \bf{24.4} & 44.6 & 52.3 & 37.1 & 58.3 & \bf{39.9} & 18.4 & 39.8 & 51.9 \\
7 & & Directly distilled by Teacher IV & 41.0 & 61.6 & 45.0 & 23.5 & 44.5 & 54.0 & 37.0 & 58.5 & 39.8 & 17.5 & 39.9 & 51.3 \\
8 & & Distilled by ensemble V+IV & 39.8 & 60.3 & 43.4 & 22.1 & 43.3 & 52.9 & 35.9 & 57.1 & 38.1 & 18.3 & 39.0 & 49.8 \\
9 & & {\bf MTPD: Teachers V$\rightarrow$IV} & \bf{41.4} & \bf{61.9} & \bf{45.1} & 23.3 & \bf{45.0} & \bf{55.4} & \bf{37.3} & \bf{58.8} & 39.8 & \bf{19.4} & \bf{40.4} & \bf{52.1} \\
\bottomrule
\end{tabular}}
\label{tab:homogenous}
\end{table*}

\subsection{Searching for Near-Optimal Teacher Orders}
\label{subsec:teacher-order}
As we have discussed in Section~\ref{subsec:multiple}, finding the optimal order of teachers for MTPD takes factorial time complexity. To acquire a near-optimal teacher order, we propose the heuristic algorithm Backward Greedy Selection (BGS, pseudo-code shown in Algorithm~\ref{alg:teacher-order}). In this section, we validate that BGS is near-optimal. To achieve this comprehensive comparison, we distill Student I with \emph{all orders} of teachers from the pool of Teachers {I-IV}. We use a reduced training budget: For each teacher, we only train the student for $3$ epochs on MS COCO. We use the linear learning rate schedule, which has been shown comparably effective in a limited budget setting by \citet{li2019budgeted}.

We first measure the adaptation costs among the student and teacher models. A visualization of the cost graph is shown in Figure~\ref{fig:dist}. Following BGS, we can construct a sequence of teachers. We compare the teacher orders given by BGS against \emph{all other} orders, via the distilled students' performance. As shown in Table~\ref{tab:teacher-order}, teacher orders suggested by BGS are consistently near-optimal in this setting. In the following sections, we use the order provided by BGS, without brute-force iterating over all possible orders.
One might argue that the greedy path selection of BGS, as shown in Figure~\ref{fig:dist}, is inferior to a global optimization algorithm.
However, we find that BGS consistently outperforms other heuristics including global optimization algorithms (see details in Appendix~\ref{sec:supp-teacher-order}). In fact, the later teachers impact the student performance more profoundly, so we need to greedily select teachers from the sequence tail.

\subsection{Distillation with Homogeneous Teachers}
\label{subsec:res50}

We start by distilling RetinaNet and Mask R-CNN with a ResNet-50 backbone (Students I \& II). Here we consider {\em homogeneous} teachers where the numbers of channels and the spatial resolutions of feature maps are {\em consistent} between the student and teacher.
For the RetinaNet student, we still consider the pool of Teachers I-IV, the same as Section~\ref{subsec:teacher-order}. For the Mask R-CNN student, we should no longer use Teacher I (the student itself) or Teacher II (the single-stage teacher does not outperform the student by a large margin). To compensate for that, we include Teacher V, which can be considered as a hybrid model of the DetectoRS backbone/neck and Mask R-CNN head. Thus, the teacher pool for Mask R-CNN includes Teachers III-V.
To control the total training time, we limit the number of teachers to be $2$. We initialize from an off-the-shelf (`OTS') student, and sequentially distill it using $2$ teachers, each with a $1\times$ training schedule. In total, the student is distilled for $24$ epochs, and the training time is equivalent to a $2\times$ training schedule. In addition to the OTS student, we also compare with three other baselines: 1) the student trained with a longer $3\times$ training schedule, which is commonly supported in object detection libraries and stronger than $1\times,2\times$ training; 2) the student \emph{directly distilled} by the final target teacher, using a $2\times$ training schedule; and 3) the student distilled by the \emph{ensemble} of teachers' feature maps.
Detector details are listed in Table~\ref{tab:stu-tea}.

Following Section~\ref{subsec:teacher-order}, we use BGS to determine the sequence of teachers to use for each student. For the RetinaNet student, BGS suggests teacher sequence III$\rightarrow$IV. For the Mask R-CNN student, BGS suggests teacher sequence V$\rightarrow$IV. Table~\ref{tab:homogenous} shows the distillation results on COCO. Additional results, analysis, and ablation studies of Mask R-CNN distillation are in Appendix~\ref{sec:supp-ablation-homo}.

{\bf Overall performance:} Our distilled student models (rows 4\&9) significantly improves over the `OTS' students (rows 1\&5). The box AP of RetinaNet is improved from $36.5\%$ to $39.9\%$ ($+3.4\%$). The box AP of Mask R-CNN is improved from $38.2\%$ to $41.4\%$ ($+3.2\%$) and the mask AP of Mask R-CNN is improved from $34.7\%$ to $37.3\%$ ($+2.6\%$). After progressive distillation, our resulting Mask R-CNN detector has {\em comparable performance with HTC teacher, but much less runtime} (51ms vs. 181ms).

{\bf Comparison with baselines:}
First, the performance gain is not merely from a longer training schedule. Our distilled student models (rows 4\&9) consistently outperform original students trained with a $3\times$ schedule (rows 2\&6). Second, progressive distillation using a curriculum of teachers (rows 4\&9) is better than direct distillation from a strong teacher (rows 3\&7), even if the total training time is the same.
Additionally, we find that using a sequence of teachers (row 9), instead of their ensemble (row 8), is more effective. This shows that integrating different types of knowledge from multiple teachers is non-trivial, and our progressive approach is better than simultaneously distilling from multiple teachers.
Notably, our detection performance for large objects receives the most gain (about $6\%$ AP$_L$ improvement for both models).
We emphasize AP$_L$ because in an efficiency-centric real-world application (e.g., autonomous driving, robot navigation), detecting nearby larger objects is more crucial than others. From a realistic perspective, better AP$_L$ shows better applicability of our approach.

\subsection{Distillation with Heterogeneous Teachers}
\label{subsec:res18}

To validate that MTPD is general, we now consider a more challenging heterogeneous scenario, where students and teachers have different backbones or input resolutions. Specifically, Student III, a ResNet-18 Mask R-CNN, is distilled with ResNet-50 teachers; Student IV, a model with reduced input resolution, is distilled with teachers trained with larger input resolutions. The results are summarized in Table~\ref{tab:heterogeneous}, and additional results are included in Appendix~\ref{sec:supp-ablation-hetero}.

{\bf Heterogeneous backbones:} Student III has a ResNet-18 backbone and about half runtime as its ResNet-50 counterpart (Teacher I). We find that the proper distillation scheme for Student III is to use the sequence of (rather than ensembling) Teachers I$\rightarrow$V$\rightarrow$IV, which significantly improves Student III over the `OTS' model. The box AP of Student III is improved from $33.3\%$ to $37.0\%$ ($+3.7\%$); and especially for large objects, AP$_L$ is improved from $43.6\%$ to $50.0\%$ ($+6.4\%$).

\begin{table}[t!]
\centering
\caption{{\bf Heterogeneous distillation of COCO detectors}, where students with smaller backbones (ResNet-18 vs. ResNet-50) or input resolutions ($333\times 200$ vs. $1333\times 800$) are distilled with heterogeneous teachers, requiring an additional feature adaptor (Section~\ref{subsec:single}). We report the detection (`Box') and segmentation (`Mask') APs, and compare our distilled student with its teachers (see Table~\ref{tab:stu-tea}), the off-the-shelf (`OTS') student, and the student distilled from an ensemble of the teachers. MTPD significantly improves the `OTS' students by over $\mathbf{3\%}$ AP.}
\resizebox{\columnwidth}{!}{
    \begin{tabular}{ @{}c | l | l l |c c@{} }
    \toprule
    \multirow{2}{*}{ID} & \multirow{2}{*}{Model} & \multirow{2}{*}{Backbone} & \multirow{2}{*}{Resolution} & \multicolumn{2}{c}{AP} \\
    & & & & Box & Mask \\
    \midrule
    1 & Student III, OTS & R18 & $1\times$ & 33.3 & 30.5 \\
    2 & Student III, Teacher Ensemble & R18 & $1\times$ & 36.0 & 32.1 \\
    3 & Student III, {\bf MTPD} & R18 & $1\times$ & \bf{37.0} & \bf{33.7} \\
    \midrule
    4 & Student IV, OTS & R50 & $0.25\times$ & 25.8 & 23.0 \\
    5 & Student IV, {\bf MTPD} & R50 & $0.25\times$ & \bf{31.5} & \bf{28.2} \\
    \bottomrule
    \end{tabular}
    }
\label{tab:heterogeneous}
\vspace{-2mm}
\end{table}

{\bf Heterogeneous input resolutions:} Although inputs with varying resolutions can be fed into most object detectors without changing the architecture, the performance often degenerates when there is a resolution mismatch between training and evaluation~\citep{tan2020efficientdet,li2020towards}. If ultimately we want to apply a detector to low-resolution inputs for fast inference, it is better to use low-resolution inputs during training. On the other hand, we conjecture that teachers with high-resolution inputs may provide finer details that can assist the student. With MTPD, we investigate the improvement of a low-resolution student distilled by a sequence of teachers with high-resolution inputs. We denote the standard input resolution $1333\times 800$ as $1\times$, and a reduced resolution $333\times 200$ as $0.25\times$. We distill Student IV (with $0.25\times$ resolution) by a sequence of Teacher I variants ($0.5\times\rightarrow0.75\times\rightarrow1\times)$. From Table~\ref{tab:heterogeneous}, we can see substantial improvement brought by MTPD: The box AP is improved from $25.8\%$ to $31.5\%$ ($+5.7\%$), and the mask AP is improved from $23.0\%$ to $28.2\%$ ($+5.2\%$).

\subsection{Generalizability to State-of-the-Art Distillation Mechanisms}
\label{subsec:combination}

Our meta-level strategy of using a sequence of teachers to progressively distill a student is independent of the choice of distillation mechanism for each teacher. We have shown MTPD can boost the simple feature-matching distillation, and in this section, we will combine MTPD with \textbf{state-of-the-art distillation mechanisms for object detection} to further improve student accuracy.

{\bf Distillation protocol:} We evaluate MTPD with three most recent methods on detector distillation: CWD~\citep{shu2021channel}, FGD~\citep{yang2022focal}, and MGD~\citep{yang2022masked}. In Appendix~\ref{sec:supp-related}, we show that classification-oriented distillation is inferior to methods delicately designed for detectors. For a fair comparison, we use the \emph{same teacher-student pairs} as them: RetinaNet/ResNet-50 and RetinaNet/ResNeXt-101~\cite{lin2017focal} are the single-stage student and final teacher. RepPoints/ResNet-50 and RepPoints/ResNeXt-101~\cite{yang2019reppoints} are the two-stage, anchor-free student and final teacher. Mask R-CNN/ResNet-50 and Cascade Mask R-CNN/ResNeXt-101-DCN~\cite{he2017mask} are the two-stage, anchor-based student and final teacher. Between them, we insert one medium-capacity teacher to progressively distill the student: RetinaNet/ResNet-101 for the first pair, RepPoints/ResNet-101 for the second, and Cascade Mask R-CNN/ResNet50-DCN for the third. Also for fairness, we \emph{keep the total training epochs the same}. We set ``$1\times$'' training schedule for each teacher, so that the total training time is equivalent to ``$2\times$,'' the same as previous work.

\begin{table}[t!]
\centering
    \vspace{-2mm}
    \caption{{\bf Distillation from Transformer-based teachers~\citep{liu2021swin} to convolution-based students.} Due to the architectural difference and capacity gap, directly distilling from a stronger teacher with Swin-S backbone does not yield better students than convolution-based teachers in Figure~\ref{fig:comb}. An intermediate Swin-T teacher and \emph{progressive distillation} solve this issue without increasing training time. Compared to off-the-shelf models, our RetinaNet and Mask R-CNN students improve by $\mathbf{5.5\%}$ AP and $\mathbf{4.3\%}$ box AP, respectively.}
\resizebox{\columnwidth}{!}{
    \begin{tabular}{ @{}c | l | l | c c @{} }
    \toprule
    \multirow{2}{*}{ID} & \multirow{2}{*}{Model} & \multirow{2}{*}{Distillation} & \multicolumn{2}{c}{AP} \\
    & & & Box & Mask \\
    \midrule
    1 & \multirow{2}{60pt}{RetinaNet (Student I)} & Direct RetinaNet/Swin-S & 41.0 & - \\
    2 & & {\bf MTPD: RetinaNet/Swin-T$\rightarrow$S} & {\bf 42.0} & - \\
    \midrule
    3 & \multirow{2}{60pt}{Mask R-CNN (Student II)} & Direct MRCNN/Swin-S & 42.0 & 37.7 \\
    4 & & {\bf MTPD: MRCNN/Swin-T$\rightarrow$S} & {\bf 42.5} & {\bf 38.4} \\
    \bottomrule
    \end{tabular}
    }
    \label{tab:swin}
\vspace{-3mm}
\end{table}

\begin{figure*}[t!]
    \centering
    \begin{subfigure}
        \centering
        \includegraphics[width=0.23\linewidth]{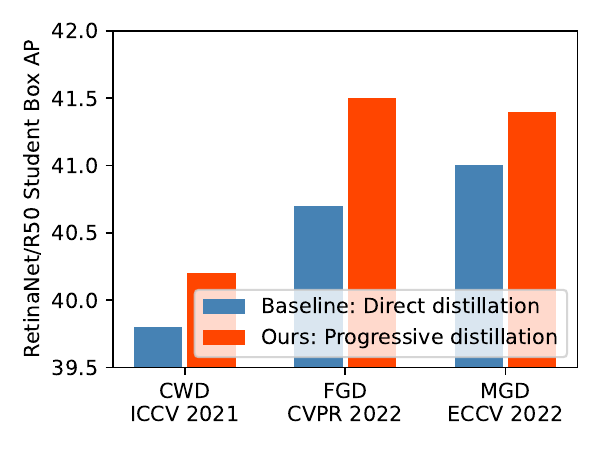}
        \label{fig:comb-a}
    \end{subfigure}
    \begin{subfigure}
        \centering
        \includegraphics[width=0.23\linewidth]{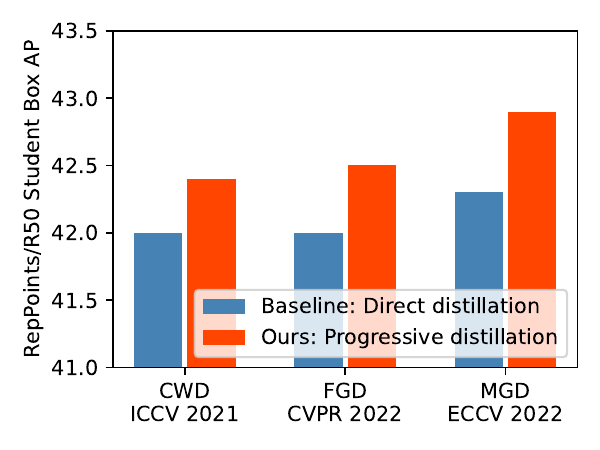}
        \label{fig:comb-d}
    \end{subfigure}
    \begin{subfigure}
        \centering
        \includegraphics[width=0.23\linewidth]{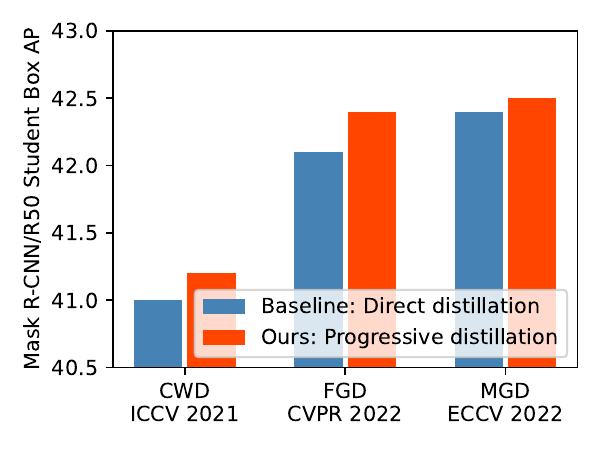}
        \label{fig:comb-b}
    \end{subfigure}
    \begin{subfigure}
        \centering
        \includegraphics[width=0.23\linewidth]{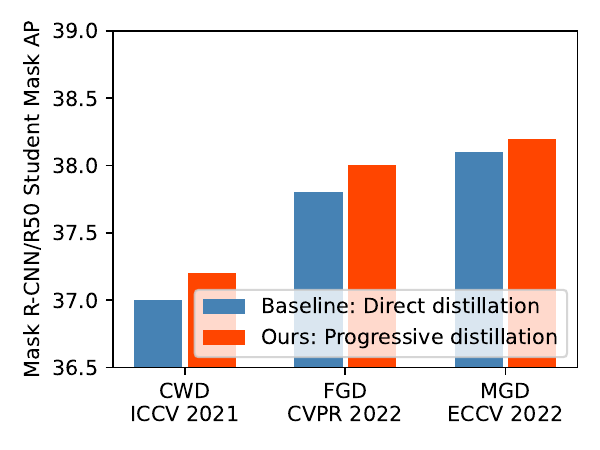}
        \label{fig:comb-c}
    \end{subfigure}\\
    \textbf{(a)\hspace{1.5in}(b)\hspace{1.5in}(c)\hspace{1.5in}(d)}
    \caption{{\bf MTPD consistently benefits state-of-the-art distillation mechanisms.} Using an intermediate RetinaNet/ResNet-101 teacher between RetinaNet/ResNet-50 student and RetinaNet/ResNeXt-101 teacher ({\bf a}), RepPoints/ResNet-101 teacher between RepPoints/ResNet-50 student and RepPoints/ResNeXt-101 teacher ({\bf b}), or Cascade Mask R-CNN/ResNet50-DCN teacher between Mask R-CNN/ResNet-50 student and Cascade Mask R-CNN/ResNeXt-101-DCN teacher ({\bf c} for Box AP and {\bf d} for Mask AP), we improve the direct distillation baselines by $0.2\%$ to $0.8\%$ AP, \emph{without increasing training time}.}
    \label{fig:comb}
\end{figure*}

Figure~\ref{fig:comb} shows that MTPD consistently improves students' final accuracy. For example, the performance of FGD-distilled RetinaNet/ResNet-50 improves from $40.7\%$ to $41.5\%$ AP ($+0.8\%$), and this gain is larger than mechanism advance from FGD to MGD ($+0.3\%$). We bring performance gains to state-of-the-art detection distillation almost \emph{for free}.

Next, we investigate how to further maximize the student performance. Due to better computation efficiency, a convolution-based (rather than Transformer-based) student is preferred. Meanwhile, Swin Transformer~\citep{liu2021swin} can act as an even stronger teacher than the convolution-based teachers used in previous work. However, compared with convolution-based teachers, direct distillation from such a teacher cannot improve the student performance, even if we use the state-of-the-art method MGD. For example, RetinaNet/Swin-Small ($47.1\%$ AP) is much stronger than RetinaNet/ResNeXt-101($41.6\%$ AP), but direct distillation from both yields the same student performance ($41.0\%$ AP). To bridge the architectural difference and capacity gap between the ResNet-50 student and Swin-Small teacher, we can utilize an intermediate Swin-Tiny teacher. As shown in Table~\ref{tab:swin}, MTPD brings the best students: the performance of ResNet-50 based RetinaNet increases to $42.0\%$ AP, and Mask R-CNN increases to $42.5\%$ AP. We also successfully distill a Transformer-based student from convolution-based teachers in Appendix~\ref{sec:supp-comb}.

\subsection{Unpacking the Performance Gain: Generalization or Optimization?}
\label{subsec:generalization}

\begin{figure*}[t!]
    \centering
    \begin{subfigure}
        \centering
        \includegraphics[width=0.33\linewidth]{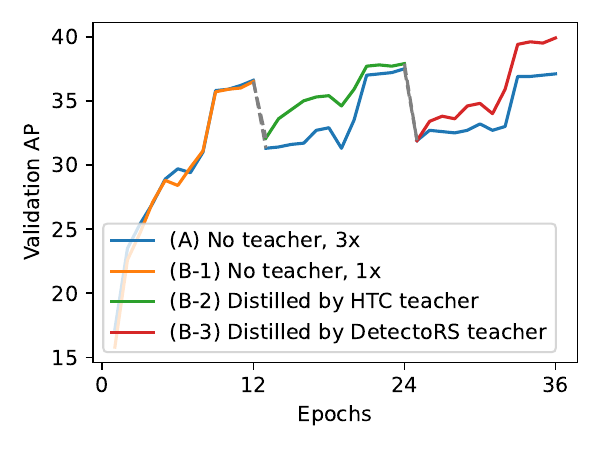}
        \label{fig:loss-a}
    \end{subfigure}
    \begin{subfigure}
        \centering
        \includegraphics[width=0.33\linewidth]{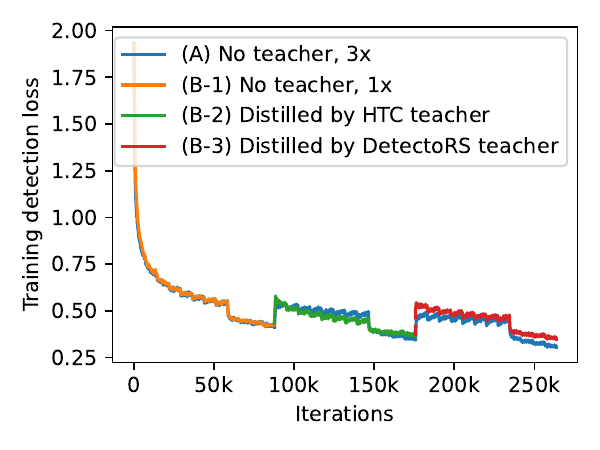}
        \label{fig:loss-b}
    \end{subfigure}
    \\
    \textbf{(a)\hspace{2.1in}(b)}
    \\
    \begin{subfigure}
        \centering
        \includegraphics[width=0.33\linewidth]{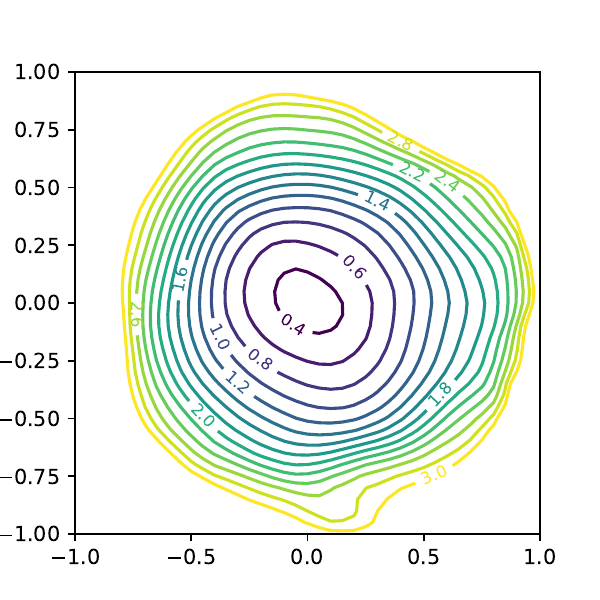}
        \label{fig:loss-c}
    \end{subfigure}
    \begin{subfigure}
        \centering
        \includegraphics[width=0.33\linewidth]{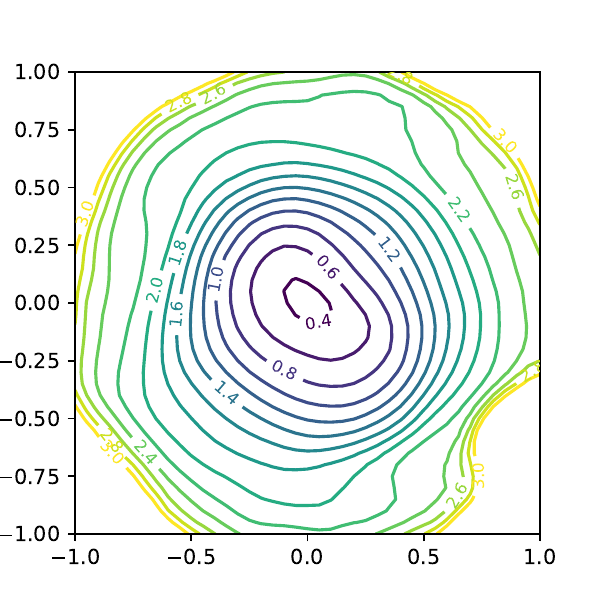}
        \label{fig:loss-d}
    \end{subfigure}
    \\
    \textbf{(c)\hspace{2.1in}(d)}
    \caption{{\bf Comparison of student models trained with and without teacher distillation.} We train a ResNet-50 backboned RetinaNet (Student I) with (A) a prolonged $3\times$ training schedule (curves in \textcolor{MidnightBlue}{blue}), or (B) MTPD from HTC (Teacher III) and then DetectoRS (Teacher IV) (curves in \textcolor{YellowOrange}{orange}-\textcolor{PineGreen}{green}-\textcolor{BrickRed}{red}). We compare the validation AP ({\bf a}) and the training detection loss $L_\text{detect}$ ({\bf b}) of the two students during the training process. Despite its worse training loss, the distilled student can generalize better on the validation set. We also compare the loss landscapes~\citep{li2017visualizing} of the original student ({\bf c}) and the distilled student ({\bf d}). Distillation can guide the student to converge to a flatter local minimum. These observations suggest that distillation \emph{helps generalization rather than optimization}.}
    \label{fig:loss}
\end{figure*}

We have shown that our distilled students significantly improve the accuracy on the {\em validation} data over off-the-shelf students. As further demonstrated in Figure~{\ref{fig:loss}a}, the validation accuracy of the distilled student gradually increases during distillation, and achieves a higher value compared with the student trained without teachers. A natural question then arises -- why is distillation helping? There are two possible hypotheses: (1) {\em improved optimization:} distillation facilitates the optimization procedure, leading to a better local minimum; and (2) {\em improved generalization:} distillation helps the student generalize to unseen data.

Improved optimization is typically manifested through a better model, a lower training loss, and a higher validation accuracy, which is exactly the case for Mask R-CNN, HTC, and DetectoRS. Consequently, one might think that distillation works in the same way. However, our investigation suggests the opposite -- MTPD increases both the validation accuracy and the training loss, and therefore effectively reduces the generalization gap.
In Figure~\ref{fig:loss}, we compare the original RetinaNet model and the distilled student, which have the same architecture, the same latency, and are trained on the same data, but with different supervision (only ground-truth labels vs. additional knowledge distillation). To eliminate the influence of learning rate changes, we train the original student with a $3\times$ schedule and restart the learning rate at the same time with the distilled student.
Interestingly, although distillation can improve the student's validation performance, the {\em training} detection loss of the distilled student is higher than the original student. This suggests that distillation does {\em not} help the optimization process to find a local minimum with a lower training loss, but rather strengthens the generalizability of the student model.

To further support this observation, we also visualize the local loss landscape~\citep{li2017visualizing}. The distilled student has a flatter loss landscape (Figure~{\ref{fig:loss}d}) compared with the original one (Figure~{\ref{fig:loss}c}). As widely believed in the machine learning literature, flat minima lead to better generalization~\citep{hochreiter1997flat,keskar2016large}. The observation shown in Figure~\ref{fig:loss} is illustrated for RetinaNet, but we also have a similar observation in other students. As a conclusion, knowledge distillation, which enforces the student to mimic the teachers' features, can be considered as an implicit regularization, and helps the student combat overfitting and achieve better generalization.

\section{Conclusion}
\label{sec:conclusion}
We present a simple yet effective approach to knowledge distillation, which progressively transfers the knowledge of a sequence of teachers to learn a lightweight object detector. Our approach automatically arranges multiple teachers into a curriculum, thus effectively mitigating the capacity gap between the teacher and student. We successfully distill knowledge from Transformer-based teachers to convolution-based students, and achieve state-of-the-art performance on the challenging COCO dataset. We also find that distillation improves generalization rather than optimization.

\noindent\textbf{Limitation and future work:} This work has mainly focused on empirical results and analysis. Due to the complexity of the detection task and models, we have not included theoretical understanding of the representation-based adaptation cost and better generalization resulted by distillation, but they will be our future direction. As a general approach to object detection, this work shares similar concerns with other detection techniques, such as potential misuse in enhancing surveillance systems, infringing upon privacy rights, or contributing to biased outcomes.

\noindent\textbf{Acknowledgement:} This work was supported in part by NSF Grant 2106825, NIFA Award 2020-67021-32799, the Jump ARCHES endowment, the NCSA Fellows program, the IBM-Illinois Discovery Accelerator Institute, the Illinois-Insper Partnership, and the Amazon Research Award. This work used NVIDIA GPUs at NCSA Delta through allocation CIS220014 from the ACCESS program.


\bibliography{reference}
\bibliographystyle{icml2023}

\newpage
\appendix
\onecolumn
\section*{Appendix}
\label{sec:appendix}
\appendix
In this appendix, we include additional contents to discuss the experimental results and details of our proposed approach, Multi-Teacher Progressive Distillation (MTPD).
Section~\ref{sec:supp-teacher-order} includes additional results and analysis of our proposed algorithm for teacher order selection. Sections~\ref{sec:supp-ablation-homo} and~\ref{sec:supp-ablation-hetero} provide additional ablation study on distillation with homogeneous teachers and heterogeneous teachers, respectively.
Section~\ref{sec:supp-other-datasets} shows the generalizability of MTPD, using demonstrate the experimental results on the Argoverse-HD dataset with streaming accuracy metric.
Section~\ref{sec:supp-related} compares MTPD with prior knowledge distillation methods in detail. Section~\ref{sec:supp-comb} provides some additional experiments on generalizing MTPD to state-of-the-art distillation mechanisms.
Section~\ref{sec:supp-impl} lists our implementation details.

\section{Additional Results on Searching for Near-Optimal Teacher Orders}
\label{sec:supp-teacher-order}

In this section, we show more detailed results about searching a proper teacher order for progressive knowledge distillation, and validate the Backward Greedy Selection (BGS) algorithm we propose in the main paper. As described in Section~\ref{subsec:multiple}, we first quantify the adaptation cost $\mathcal{C}(\cdot,\cdot)$ between every pair of models in our pool, and then use a heuristic method BGS (Algorithm~\ref{alg:teacher-order}) to construct a sequence of teachers. We have shown that the teacher order suggested by BGS is highly competitive in Table~\ref{tab:teacher-order}. One might think that there should be better choices than a greedy algorithm on a directed graph, such as a shortest-path algorithm. To validate our algorithm design, we empirically compare BGS against several other algorithms.

\begin{center}
\noindent\resizebox{0.95\columnwidth}{!}{
\begin{minipage}{\columnwidth}
\begin{algorithm}[H]
    \SetAlgoLined
    \KwInput{Trained student model $S$, pool of trained teacher models $\mathcal{P}=\{T_i\}_{i=1}^N$, teacher models' performance $\{Q(T_i)\}_{i=1}^N$, maximum number of selected teachers $k$}
    \KwOutput{Sequence of teachers $\mathcal{O}, \text{len}(\mathcal{O})\le k$}
    Pick the best performing teacher: $T_{\alpha_k}\gets\arg\max_{T_u\in \mathcal{P}}Q(T_u)$, $\mathcal{O}\gets[T_{\alpha_k}]$ \\
    Exclude from pool: $\mathcal{P}\gets \mathcal{P} \setminus \{T_{\alpha_k}\}$ \\
    \For{$j\gets k-1$ \KwTo $1$}
    {
        Get candidate sub-pool: $\mathcal{P}_j=\{T_u\mid T_u\in \mathcal{P}, \mathcal{C}(T_u,T_{\alpha_{j+1}})<\mathcal{C}(S,T_{\alpha_{j+1}})\}$ \\
        \If{$\mathcal{P}_j\ne\emptyset$}
        {
            Pick the teacher closest to $T_{\alpha_{j+1}}$: $T_{\alpha_j}\gets\arg\min_{T_u\in \mathcal{P}_j}\mathcal{C}(T_u, T_{\alpha_{j+1}})$ \\
            Prepend $T_{\alpha_j}$ to $\mathcal{O}$ \\
            Exclude from pool: $\mathcal{P}\gets \mathcal{P} \setminus \{T_{\alpha_j}\}$
        }
        \lElse*{Break}
    }
    \Return{$\mathcal{O}$}
    \caption{Backward Greedy Selection (BGS) for determining the teacher order.}
    \label{alg:teacher-order}
\end{algorithm}
\end{minipage}}
\end{center}

To begin with, we include the detailed adaptation costs $\mathcal{C}(\cdot,\cdot)$ among RetinaNet (Student I) and its teachers (Teachers I-IV) in Table~\ref{tab:cost}. As described in Section~\ref{subsec:teacher-order}, we have distilled Student I with {\em all} possible teacher orders in the pool, using a reduced training budget of 3 epochs for each teacher. The results of these mini-budget distillation are summarized in Table~\ref{tab:mini}.

\begin{table*}[h!]
    \centering
    \caption{Adaptation costs among Student I (RetinaNet) and Teachers I-IV (Mask R-CNN, FCOS, HTC, DetectoRS). The adaptation cost is computed pair-wise as described in Section~\ref{subsec:multiple} of the main paper.}
    \begin{tabular}{l|c c c c c}
    \toprule
    \backslashbox{From}{To} & Student I & Teacher I & Teacher II & Teacher III & Teacher IV \\
    \midrule
    Student I & - & 0.939 & 0.060 & 1.568 & 1.254 \\
    Teacher I & 0.183 & - & 0.070 & 0.934 & 0.963 \\
    Teacher II & 0.339 & 1.181 & - & 1.940 & 1.401 \\
    Teacher III & 0.191 & 0.484 & 0.082 & - & 0.890 \\
    Teacher IV & 0.232 & 0.767 & 0.077 & 1.248 & - \\
    \bottomrule
    \end{tabular}
    \label{tab:cost}
\end{table*}

\begin{table*}[h!]
    \centering
    \caption{Performance of Student I (RetinaNet) distilled with different teacher sequences, under reduced training budgets. For each teacher in the sequence, the student is trained for $3$ epochs on COCO. After progressive knowledge distillation, the student is evaluated on the COCO validation set. The teacher orders suggested by BGS are marked {\bf bold}. No teacher order is highlighted in $k=4$, as BGS determines that the adaptation cost from Teacher II to Teacher I is already higher than that from the student to Teacher I, so only three teachers (I$\rightarrow$III$\rightarrow$IV) should be used. }
    \begin{tabular}{c|l c||c|l c||c|l c}
    \toprule
    \multirow{2}{*}{Length} & Teacher & Student & \multirow{2}{*}{Length} & Teacher & Student & \multirow{2}{*}{Length} & Teacher & Student \\
     & Sequence & AP & & Sequence & AP & & Sequence & AP \\
    \midrule
\multirow{4}{*}{1} & III & 36.8 & \multirow{24}{*}{3} & III$\rightarrow$II$\rightarrow$IV & 38.0 & \multirow{24}{*}{4} & III$\rightarrow$II$\rightarrow$I$\rightarrow$IV & 38.2 \\
  & \bf{IV} & \bf{36.7} &   & {\bf I$\rightarrow$III$\rightarrow$IV} & {\bf 37.9} &   & III$\rightarrow$I$\rightarrow$II$\rightarrow$IV & 38.1 \\
  & I & 36.4 &   & III$\rightarrow$IV$\rightarrow$II & 37.9 &   & I$\rightarrow$III$\rightarrow$II$\rightarrow$IV & 38.1 \\
  & II & 36.2 &   & II$\rightarrow$III$\rightarrow$IV & 37.9 &   & II$\rightarrow$III$\rightarrow$I$\rightarrow$IV & 38.0 \\
  &   &   &   & III$\rightarrow$I$\rightarrow$IV & 37.8 &   & I$\rightarrow$III$\rightarrow$IV$\rightarrow$II & 38.0 \\
\multirow{12}{*}{2} & {\bf III$\rightarrow$IV} & {\bf 37.6} &   & I$\rightarrow$II$\rightarrow$IV & 37.7 &   & II$\rightarrow$I$\rightarrow$III$\rightarrow$IV & 37.9 \\
  & IV$\rightarrow$II & 37.3 &   & I$\rightarrow$IV$\rightarrow$II & 37.6 &   & III$\rightarrow$I$\rightarrow$IV$\rightarrow$II & 37.9 \\
  & III$\rightarrow$II & 37.3 &   & IV$\rightarrow$II$\rightarrow$III & 37.5 &   & I$\rightarrow$II$\rightarrow$III$\rightarrow$IV & 37.9 \\
  & I$\rightarrow$IV & 37.3 &   & IV$\rightarrow$III$\rightarrow$II & 37.5 &   & IV$\rightarrow$I$\rightarrow$III$\rightarrow$II & 37.7 \\
  & IV$\rightarrow$III & 37.2 &   & II$\rightarrow$I$\rightarrow$IV & 37.5 &   & II$\rightarrow$I$\rightarrow$IV$\rightarrow$III & 37.7 \\
  & I$\rightarrow$III & 37.1 &   & I$\rightarrow$III$\rightarrow$II & 37.5 &   & I$\rightarrow$II$\rightarrow$IV$\rightarrow$III & 37.7 \\
  & IV$\rightarrow$I & 37.0 &   & IV$\rightarrow$I$\rightarrow$III & 37.4 &   & IV$\rightarrow$III$\rightarrow$I$\rightarrow$II & 37.6 \\
  & II$\rightarrow$IV & 37.0 &   & II$\rightarrow$IV$\rightarrow$III & 37.4 &   & IV$\rightarrow$I$\rightarrow$II$\rightarrow$III & 37.6 \\
  & III$\rightarrow$I & 37.0 &   & III$\rightarrow$IV$\rightarrow$I & 37.4 &   & III$\rightarrow$IV$\rightarrow$II$\rightarrow$I & 37.6 \\
  & II$\rightarrow$I & 36.9 &   & III$\rightarrow$I$\rightarrow$II & 37.4 &   & III$\rightarrow$IV$\rightarrow$I$\rightarrow$II & 37.6 \\
  & II$\rightarrow$III & 36.8 &   & I$\rightarrow$IV$\rightarrow$III & 37.4 &   & III$\rightarrow$II$\rightarrow$IV$\rightarrow$I & 37.6 \\
  & I$\rightarrow$II & 36.8 &   & IV$\rightarrow$II$\rightarrow$I & 37.3 &   & I$\rightarrow$IV$\rightarrow$III$\rightarrow$II & 37.6 \\
  &   &   &   & IV$\rightarrow$III$\rightarrow$I & 37.3 &   & IV$\rightarrow$II$\rightarrow$I$\rightarrow$III & 37.5 \\
  &   &   &   & IV$\rightarrow$I$\rightarrow$II & 37.3 &   & IV$\rightarrow$III$\rightarrow$II$\rightarrow$I & 37.5 \\
  &   &   &   & I$\rightarrow$II$\rightarrow$III & 37.3 &   & II$\rightarrow$IV$\rightarrow$I$\rightarrow$III & 37.5 \\
  &   &   &   & II$\rightarrow$IV$\rightarrow$I & 37.2 &   & II$\rightarrow$III$\rightarrow$IV$\rightarrow$I & 37.5 \\
  &   &   &   & II$\rightarrow$I$\rightarrow$III & 37.2 &   & I$\rightarrow$IV$\rightarrow$II$\rightarrow$III & 37.5 \\
  &   &   &   & III$\rightarrow$II$\rightarrow$I & 37.2 &   & II$\rightarrow$IV$\rightarrow$III$\rightarrow$I & 37.4 \\
  &   &   &   & II$\rightarrow$III$\rightarrow$I & 37.1 &   & IV$\rightarrow$II$\rightarrow$III$\rightarrow$I & 37.3 \\
    \bottomrule
    \end{tabular}
    \label{tab:mini}
\end{table*}

\begin{table*}[h!]
    \centering
\caption{Comparison of five heuristic algorithms for teacher order selection in the mini-budget and full-budget distillation settings. Our BGS can consistently produce the best teacher order among all candidate algorithms.}
    \resizebox{\columnwidth}{!}{
    \begin{tabular}{c|l|l|c c|c||c|l|l|c c|c}
        \toprule
        \multirow{2}{*}{$k$} & \multirow{2}{*}{Algorithm} & Teacher & \multicolumn{2}{c|}{Mini-Budget} & Full-Budget & \multirow{2}{*}{$k$} & \multirow{2}{*}{Algorithm} & Teacher & \multicolumn{2}{c|}{Mini-Budget} & Full-Budget \\
        & & Order & Student AP & Ranking & Student AP & & & Order & Student AP & Ranking & Student AP \\
        \midrule
        \multirow{5}{*}{1} & Shortest-path (sum) & IV & 36.7 & 2 / 4 & 39.2 & \multirow{5}{*}{3} & Shortest-path (sum) & II$\rightarrow$I$\rightarrow$IV & 37.5 & 9 / 40 & 39.3 \\
         & Shortest-path (max) & IV & 36.7 & 2 / 4 & 39.2 & & Shortest-path (max) & I$\rightarrow$III$\rightarrow$IV & 37.9 & 2 / 40 & 39.9 \\
         & Forward construction & II & 36.2 & 4 / 4 & 36.8 & & Forward construction & II$\rightarrow$I$\rightarrow$III & 37.2 & 25 / 40 & 37.8 \\
         & Top-$k$ performance & IV & 36.7 & 2 / 4 & 39.2 & & Top-$k$ performance & II$\rightarrow$III$\rightarrow$IV & 37.9 & 2 / 40 & 39.7 \\
         & BGS (Ours) & IV & 36.7 & 2 / 4 & 39.2 & & BGS (Ours) & I$\rightarrow$III$\rightarrow$IV & 37.9 & 2 / 40 & 39.9 \\
        \midrule
        \multirow{5}{*}{2} & Shortest-path (sum) & II$\rightarrow$IV & 37.0 & 7 / 16 & 39.3 & \multirow{5}{*}{4} & Shortest-path (sum) & II$\rightarrow$I$\rightarrow$III$\rightarrow$IV & 37.9 & 7 / 64 & 39.3 \\
         & Shortest-path (max) & I$\rightarrow$IV & 37.3 & 2 / 16 & 39.5 & & Shortest-path (max) & II$\rightarrow$I$\rightarrow$III$\rightarrow$IV & 37.9 & 7 / 64 & 39.3 \\
         & Forward construction & II$\rightarrow$I & 36.9 & 10 / 16 & 37.5 & & Forward construction & II$\rightarrow$I$\rightarrow$III$\rightarrow$IV & 37.9 & 7 / 64 & 39.3 \\
         & Top-$k$ performance & III$\rightarrow$IV & 37.6 & 1 / 16 & 39.9 & & Top-$k$ performance & I$\rightarrow$II$\rightarrow$III$\rightarrow$IV & 37.9 & 7 / 64 & 39.7 \\
         & BGS (Ours) & III$\rightarrow$IV & 37.6 & 1 / 16 & 39.9 & & BGS (Ours) & I$\rightarrow$III$\rightarrow$IV & 37.9 & 7 / 64 & 39.9 \\
        \bottomrule
    \end{tabular}}
\label{tab:alg}
\end{table*}

\clearpage

Given the adaptation costs in Table~\ref{tab:cost}, we can construct a directed graph, part of which has been illustrated in Figure~\ref{fig:dist}. On the directed graph, we run several algorithms to select a path. In addition to BGS, one may also propose these algorithms:
\begin{itemize}[leftmargin=*, noitemsep, nolistsep]
    \item {\bf {Shortest-path (sum)}}: Set the student as the source node, and set the best performing teacher as the target node $T_{\lambda_k}$. Find a path $S\rightarrow T_{\lambda_1}\rightarrow\dots\rightarrow T_{\lambda_k}$ that minimizes the {\em sum} of adaptation costs along the path:
    
    $\min_{T_{\lambda_1},\dots,T_{\lambda_{k-1}}}\mathcal{C}(S,T_{\lambda_1})+\sum_{j=1}^{k-1}\mathcal{C}(T_{\lambda_j},T_{\lambda_{j+1}})$.
    \item {\bf {Shortest-path (max)}}: Set the student as the source node, and set the best performing teacher as the target node $T_{\lambda_k}$. Find a path $S\rightarrow T_{\lambda_1}\rightarrow\dots\rightarrow T_{\lambda_k}$ that minimizes the {\em maximum} of adaptation costs along the path: $\min_{T_{\lambda_1},\dots,T_{\lambda_{k-1}}}\max\{\mathcal{C}(S,T_{\lambda_1}),\mathcal{C}(T_{\lambda_1},T_{\lambda_2}),\dots,\mathcal{C}(T_{\lambda_{k-1}},T_{\lambda_k})\}$.
    \item {\bf {Forward construction}}: Contrary to BGS, we may start from the student and choose the nearest teacher from the current one, to construct the sequence: $T_{\lambda_1}\gets\arg\min_{T_u\in\mathcal{P}}\mathcal{C}(S,T_u),T_{\lambda_{j+1}}\gets\arg\min_{T_u\in\mathcal{P}}\mathcal{C}(T_{\lambda_j},T_u)$.
    \item {\bf {Top-$k$ performance}}: Find the best performing top-$k$ teachers which satisfy $Q(T_{\lambda_1})\le\dots\le Q(T_{\lambda_k})$, and set the path as $S\rightarrow T_{\lambda_1}\rightarrow\dots\rightarrow T_{\lambda_k}$.
\end{itemize}

The output teacher sequences and corresponding student performance of these four algorithms are summarized in Table~\ref{tab:alg}. In this setting, BGS can consistently produce a competitive teacher order that leads to a good performance of the distilled student. The other four algorithms all propose teacher sequences that are worse than or equivalent to BGS. In other words, these additional algorithm candidates fail to outperform BGS.

In addition to the mini-budget setting, we also verify that the superiority of BGS holds when using the full training schedule. To this end, we perform progressive distillation with teacher orders suggested by various algorithms, this time using a full-budget training schedule (12 epochs for each teacher). Due to limited computation resources, we cannot extensively evaluate all possible teacher orders with full budgets, preventing us from obtaining their absolute ranking among all possibilities. Nevertheless, we can still compare the teacher orders produced by different algorithms. The results are presented in the last column in Table~\ref{tab:alg}. Again, the teacher orders generated by BGS consistently outperform those from all other heuristic algorithms when using a full distillation schedule.

We note that the top-$k$ performance algorithm is the best among the four candidates and is comparable to BGS. Though different teacher sequences are selected, they would lead to the same final student performance as BGS in the mini-budget setting. Meanwhile, the full-budget results show that BGS outperforms the top-$k$ performance algorithm. We believe that such a performance ranking based algorithm is worse or on par with BGS when the teachers have similar architectures (e.g., all using ResNet backbones).

However, when we include more diverse teacher models in the pool, the top-$k$ performance algorithm encounters a pitfall. For example, in the experiment (see Table~\ref{tab:swin}) where we distill Mask R-CNN students with both Transformer-based and convolution-based teachers in the pool, the top-$k$ performance algorithm would suggest Cascade MRCNN/ResNeXt-101-DCN$\rightarrow$Mask R-CNN/Swin-S, leading to a final student performance of 42.0 AP. In contrast, BGS suggests Mask R-CNN/Swin-T$\rightarrow$Mask R-CNN/Swin-S, achieving a higher student performance of 42.5 AP.
This difference highlights that merely relying on performance ranking can result in an ineffective curriculum of teachers, as it does not account for the representation similarity between student and teacher models during distillation. In a successful distillation process, it is crucial to have a degree of similarity between the student and teacher models to facilitate the transfer of knowledge. Our similarity-based teacher sequence design, which incorporates feature-level measurements, ensures a more effective learning process for the student. Consequently, BGS proves to be more effective than algorithms that solely focus on performance ranking.

In summary, our greedy backward construction, BGS, works the best in our setting, outperforming globally optimized shortest-path algorithms or a performance ranking based algorithm. The final target teacher has the most profound impact on the distilled student's performance. In order to fully assist the final teacher, we need to use another teacher with the minimal adaptation cost to the final teacher before it, which is exactly the behavior of BGS.

\section{Ablation Study on Distillation with Homogeneous Teachers}
\label{sec:supp-ablation-homo}

In this section, we provide more details about distillation with homogeneous teachers (Section~\ref{subsec:res50}). We investigate (1) the impact of each individual teacher; and (2) distillation with teachers simultaneously vs. sequentially.

\noindent{\bf Impact of individual teachers:}
We first distill Student II with each of the three teachers individually: Teacher III has the same backbone and neck but a more advanced head; Teacher IV has more advanced backbone, neck, and head; Teacher V has the same head but more advanced backbone and neck.
Table~\ref{tab:res50-indivisual} provides the performance of the three teachers, where Teacher IV achieves the best performance (rows 1-3). From Table~\ref{tab:res50-combine}, we can see that our distilled students (rows 2-7) {\em significantly and consistently} outperform the off-the-shelf student (row 1), demonstrating the effectiveness of our MTPD strategy {\em irrespective of the types of teachers}. Moreover, the improvement of the student distilled with Teacher V (row 2) over that with Teacher III (row 3) shows that a more powerful teacher generally leads to a better distilled student. Interestingly, although Teacher IV is more powerful than Teacher V, Table~\ref{tab:res50-combine} shows that their distilled students achieve quite similar AP (row 2 vs. row 4). This indicates that an even more powerful teacher does not necessarily further improve the distilled student; too large a capacity and structure gap between the teacher and student will limit the effectiveness of distillation. Also, it is easier to distill from teachers with the same head.

\noindent{\bf Simultaneous vs. progressive distillation:}
We now distill Student II with the combination of teachers, and we choose the top-performing Teacher IV and Teacher V. We investigate two types of combination -- simultaneous distillation with a feature matching loss between the student and the ensemble of the teachers (row 5), and sequential distillation with teachers one by one (rows 6-7). First, we find that using both teachers simultaneously (row 5) is {\em even worse} than using a single teacher (rows 2-4). This shows that integrating different types of knowledge from multiple teachers is not a trivial task -- simultaneously using the features from multiple teachers might provide {\em conflicting supervisions} to the student model and thus hinder its distillation process. By contrast, our sequential distillation overcomes this issue and improves the performance {\em irrespective of the order of the teachers} (rows 6-7 vs. rows 1-4). Second, the sequential order of the teachers makes a difference. A {\em curriculum-like progression} (row 7),
where the teacher with a smaller adaptation cost is used first and that with a larger adaptation cost and a higher performance is used later, leads to the best performance.

\begin{table*}[t!]
    \centering
\caption{Homogeneous distillation of COCO detectors, where students with ResNet-50 backbones are distilled with teachers with ResNet-50 backbones. We report the detection (`Box') and segmentation (`Mask') APs and runtime, and we compare our distilled student with its teachers, and off-the-shelf (`OTS') student. Our distilled student significantly improves the APs over the `OTS' student by around $3\%$.}
    \resizebox{\columnwidth}{!}{
        \begin{tabular}{ @{}c | l|c c c c c c|c c c c c c|c @{} }
            \toprule
          \multirow{2}{*}{ID}& \multirow{2}{*}{Model} & \multicolumn{6}{c|}{Box} & \multicolumn{6}{c|}{Mask} & Runtime \\
             & & AP & AP$_{50}$ & AP$_{75}$ & AP$_S$ & AP$_M$ & AP$_L$ & AP & AP$_{50}$ & AP$_{75}$ & AP$_S$ & AP$_M$ & AP$_L$ & (ms)\\
    \midrule
    1 & Teacher III & 42.3 & 61.1 & 45.8& 23.7& 45.6 & 56.3& 37.4& 58.4 & 40.2 & 19.6 & 40.4 & 51.7 & 181 \\
    2 & Teacher IV &  49.1& 67.7 &  53.4 & 29.9& 53.0& 65.2& 42.6& 65.1 & 46.0 & 24.1 & 46.4 & 58.6 & 223 \\
    3 & Teacher V &  45.1 & 66.3 & 49.3 & 27.8 & 49.0& 59.3 & 40.1 & 63.1 & 42.8 & 22.9 & 43.8 & 54.8 & 142 \\
    \midrule 
    4 & Student II (OTS) & 38.2 & 58.8 & 41.4& 21.9& 40.9 & 49.5 & 34.7 & 55.7& 37.2& 18.3& 37.4& 47.2& 51 \\
    5 & Student II (Ours) & \bf{41.4} &  \bf{61.9} & \bf{45.1}& \bf{23.3}& \bf{45.0}& \bf{55.4}&\bf{37.3} & \bf{58.8}& \bf{39.8}& \bf{19.4}& \bf{40.4}& \bf{52.1}& 49 \\
    \bottomrule
\end{tabular}}
\label{tab:res50-indivisual}
\end{table*}

\begin{table*}
    \centering
\caption{Ablation study on homogeneous distillation of COCO detectors (models in Table~\ref{tab:res50-indivisual}).  
Our distillation strategy is {\em consistently effective irrespective of the types of teachers}. Moreover, sequential distillation with two teachers outperforms both distillation with a single teacher and simultaneous distillation with two teachers. Our best distilled student is obtained by MTPD, where Student II is first distilled with Teacher V (a weaker, more similar teacher with the same head as Student II) and then distilled with Teacher IV (a stronger teacher whose architecture is completely different from Student II).}
    \resizebox{\columnwidth}{!}{
        \begin{tabular}{@{} c | l|c c c c c c|c c c c c c @{} }
            \toprule
          \multirow{2}{*}{ID} & \multirow{2}{*}{Student II} &  \multicolumn{6}{c|}{Box} & \multicolumn{6}{c}{Mask}  \\
            & & AP & AP$_{50}$ & AP$_{75}$ & AP$_S$ & AP$_M$ & AP$_L$ & AP & AP$_{50}$ & AP$_{75}$ & AP$_S$ & AP$_M$ & AP$_L$ \\
    \midrule
    1 & OTS & 38.2 & 58.8 & 41.4& 21.9& 40.9 & 49.5 & 34.7 & 55.7& 37.2& 18.3& 37.4& 47.2 \multirow{5}{*}{} \\
    \midrule
    2 &  Distilled by Teacher III & 40.2 & 60.7 & 43.8& 22.5& 43.8& 53.4& 36.3& 57.3& 38.7& 18.9& 39.3& 50.3 \\
    3 & Distilled by Teacher IV & {40.8}  & 61.5& 44.6 & 23.0 & 44.3& 54.2& {36.8}& 58.3 & 39.4& 19.2& 39.9& 51.0 \\
    4 & Distilled by Teacher V & 40.8 & 61.4 & 44.5& 22.9& 44.3& 54.2& 36.6& 58.1& 39.1& 19.2& 39.6& 51.0 \\ \midrule
     5&    Distilled by Teachers IV+V & 39.8 & 60.3& 43.4& 22.1& 43.3& 52.9& 35.9& 57.1& 38.1& 18.3& 39.0& 49.8\\ \midrule
  6 & Distilled by Teachers IV$\rightarrow$V & 41.0 & 61.7 & 44.8 & 23.0 & 44.3 & 54.9 & 36.8 & 58.3& 39.2& 19.5& 39.9& 51.3  \\ %
  7 & Distilled by Teachers V$\rightarrow$IV & \bf{41.4} &  \bf{61.9} & \bf{45.1}& \bf{23.3}& \bf{45.0}& \bf{55.4}&\bf{37.3} & \bf{58.8}& \bf{39.8}& \bf{19.4}& \bf{40.4}& \bf{52.1}\\
    \bottomrule
\end{tabular}}
\label{tab:res50-combine}
\end{table*}

\noindent{\bf Overall performance:} Our best distillation performance is achieved when we first distill Student II with a curriculum of teachers (Teachers V$\rightarrow$IV). Overall, the box AP is improved from $38.2\%$ to $41.4\%$, and the mask AP is improved from $34.7\%$ to $37.3\%$. Our resulting Mask R-CNN detector has {\em comparable performance with HTC, but much smaller runtime}.

\section{Ablation Study on Distillation with Heterogeneous Teachers}
\label{sec:supp-ablation-hetero}

In this section, we provide more details about distillation with heterogeneous teachers (Section~\ref{tab:res18}). We investigate the heterogeneous cases where the backbones or input resolutions are different between the teachers and student.

\noindent{\bf Overall performance:} Again, Tables~\ref{tab:res18} and~\ref{tab:res18-ablations} show that MTPD is consistently effective with respect to all the teachers and their combinations, e.g., the box AP improves from $33.3\%$ to $37.0\%$, and the mask AP improves from $30.5\%$ to $33.7\%$.

\noindent{\bf Two key findings in heterogeneous distillation:} Compared with the homogeneous case, we find that the capacity gap between models is a more important factor, and to bridge this gap a proper teacher order plays a more critical role. Details are explained as follows.

\noindent{\em The student-teacher capacity gap is more pronounced in heterogeneous distillation.} Among the four teachers, Teacher I shares exactly the same neck and head structure with the student, and has a similar but larger backbone; Teacher V has the same head with the student as well, but has different backbone and neck; Teacher III has similar backbone and neck, but has a different head; and Teacher IV is the most powerful one with completely different architecture. Table~\ref{tab:res18-ablations} (rows 3-6) summarizes the distillation results with single teachers. First, directly distilling from the strongest teacher (Teacher IV) does not yield the largest improvement. Second, a relatively less powerful but more similar teacher (Teacher I) leads to the best distillation performance, improving the APs by 2\%, although Teachers V, III, and IV are all stronger than Teacher I. One possible reason is that Teacher I has the same neck and head as Student III as well as a similar but deeper backbone, so the capacity gap between Student III and Teacher I is the smallest. Finally, we find that Teacher III is a strong but not particularly helpful teacher, achieving the worst distillation results. One possible reason is that Teacher III has a very different head from Student III, while not as accurate as Teacher IV, making it unable to provide enough guidance to Student III.
These observations suggest that a smaller capacity gap between the student and the teacher may facilities knowledge transfer.

\begin{table*}[t!]
    \centering
\caption{Heterogenous distillation of COCO detectors, where students with ResNet-18 backbones are distilled with teachers with ResNet-50 backbones. We report the detection (`Box') and segmentation (`Mask') APs and runtime, and we compare our distilled student with its teachers and off-the-shelf (`OTS') student. Our distilled student significantly improves the APs over the `OTS' student by over $3\%$.}
    \resizebox{\columnwidth}{!}{
\begin{tabular}{ @{}c | l|c c c c c c|c c c c c c|c @{} }
    \toprule
    \multirow{2}{*}{ID} & \multirow{2}{*}{Model} & \multicolumn{6}{c|}{Box} & \multicolumn{6}{c|}{Mask} & Runtime \\
     & &  AP & AP$_{50}$ & AP$_{75}$ & AP$_S$ & AP$_M$ & AP$_L$ & AP & AP$_{50}$ & AP$_{75}$ & AP$_S$ & AP$_M$ & AP$_L$ & (ms)\\
    \midrule
    1 & Teacher I & 38.2 & 58.8 & 41.4 & 21.9& 40.9& 49.5& 34.7& 55.7 & 37.2 & 18.3 & 37.4 & 47.2 & 51\\
    2 & Teacher III &  42.3 & 61.1 & 45.8& 23.7& 45.6 & 56.3& 37.4& 58.4 & 40.2 & 19.6 & 40.4 & 51.7 & 181 \\
    3 & Teacher IV &  49.1& 67.7 &  53.4 & 29.9& 53.0& 65.2& 42.6& 65.1 & 46.0 & 24.1 & 46.4 & 58.6 & 223\\   
    4 & Teacher V &  45.1 & 66.3 & 49.3 & 27.8 & 49.0& 59.3 & 40.1 & 63.1 & 42.8 & 22.9 & 43.8 & 54.8 & 142 \\
     \midrule 
    5 & Student III (OTS) & 33.3 & 52.9 & 35.9& 18.2 & 35.9& 43.6& 30.5& 50.0 & 32.1 & 15.5 & 32.9 & 41.8 & 29 \\
   6 & Student III (Ours) & \bf{37.0} & \bf{56.8} & \bf{39.9}& \bf{20.2}& \bf{39.8} & \bf{50.0} & \bf{33.7} &  \bf{53.6} &\bf{36.0} &\bf{17.2} &\bf{36.0} &\bf{47.3} & 29\\ 
    \bottomrule
\end{tabular}
    }
\label{tab:res18}
\end{table*}

\begin{table*}[h!]
    \centering
\caption{Ablation study on heterogeneous COCO detector distillation (models in Table~\ref{tab:res18}). Student III (Mask R-CNN with a ResNet-18 backbone) is distilled with teachers with {\em different} and larger ResNet-50 backbones. Training Student III for more epochs improves its performance, but not as much as progressive distillation with teachers. Note that for each distillation we train 12 epochs. MTPD is {\em consistently effective irrespective of the types of teachers}. Moreover, MTPD outperforms simultaneous distillation by ensembling multiple teachers. Our best distilled student is obtained by MTPD, where Student III is first distilled with Teacher I (the most similar teacher with the same head and neck as Student III and a deeper backbone), then distilled with Teacher V (the stronger teacher with the same head as Student III), and finally distilled with Teacher IV (the strongest teacher whose architecture is completely different from Student III).}
    \resizebox{\columnwidth}{!}{
\begin{tabular}{ @{} c|l|c c c c c c|c c c c c c @{} }
\toprule     
\multirow{2}{*}{ID} & \multirow{2}{*}{Model} & \multicolumn{6}{c|}{Box} & \multicolumn{6}{c}{Mask} \\
      & & AP & AP$_{50}$ & AP$_{75}$ & AP$_S$ & AP$_M$ & AP$_L$ & AP & AP$_{50}$ & AP$_{75}$ & AP$_S$ & AP$_M$ & AP$_L$ \\
    \midrule
     1 & Student III (OTS) & 33.3 & 52.9 & 35.9& 18.2 & 35.9& 43.6& 30.5& 50.0 & 32.1 & 15.5 & 32.9 & 41.8 \\
    \midrule
     2 & +12 epochs & 34.6 & 54.5 & 37.2& 18.8& 36.9& 46.1& 31.6 & 51.5 & 33.6 & 15.8 & 33.7 & 44.0  \\ 
     3 & +24 epochs & 34.5 & 54.2 & 37.2 & 18.8 & 36.5 & 45.8 & 31.5 & 51.2 & 33.8 & 16.0 & 33.4 & 43.7 \\
     4 & +36 epochs & 34.6 & 54.2 & 37.4 & 18.6 & 36.9 & 46.7 & 31.6 & 51.1 & 33.8 & 15.7 & 33.6 & 44.3 \\
     \midrule
     3 & Distilled by Teacher I & 35.8 & 55.8 & 38.8& 19.3& 38.8& 47.9 & 32.6 & 52.7 & 34.8 & 16.0& 35.3& 45.5 \\
     4 & Distilled by Teacher III & 35.2 & 55.2 & 37.8 & 19.1 & 37.8& 47.4& 32.1& 52.0& 34.0& 16.1& 34.5& 45.2 \\ 
     5 & Distilled by Teacher IV & 35.5 & 55.2 & 38.2 & 19.0 & 37.9 & 48.0 & 32.4 & 51.9 & 34.5 & 15.9 & 34.8& 45.6  \\
     6 & Distilled by Teacher V & 35.4 & 55.2 & 38.3 & 19.4 & 37.9 & 48.4 & 32.2 & 52.2 & 34.3 & 15.4& 34.4& 45.8 \\      \midrule
     7 & Distilled by Teachers IV$+$V & 34.8 & 54.9 & 37.2 & 19.0 & 37.2& 47.0& 31.6& 51.7& 33.9& 15.7& 33.8& 44.2 \\ 
     8 & Distilled by Teachers I$+$IV$+$V & 36.0 & 55.4 & 39.1& 18.2& 38.1& 48.3& 32.1& 53.0 & 34.7 & 15.8 & 34.7 & 46.1  \\
     9 & Distilled by Teachers I$+$III$+$IV+$V$ & 36.1 & 55.2 & 39.0& 18.4& 38.2& 48.0& 31.7& 52.9 & 34.3 & 15.1 & 34.2 & 46.3 \\     \midrule
    10 & Distilled by Teachers I$\rightarrow$V & 36.5 & 56.3 & 39.3& 19.5& 38.8& 49.4& 33.2& 53.2 & 35.3 & 16.4 & 35.4 & 46.8  \\
     11 & Distilled by Teachers V$\rightarrow$IV & 35.2 & 55.2 & 37.8 & 19.1 & 37.8& 47.4& 32.1& 52.0& 34.0& 16.1& 34.5& 45.2 \\
     12 & Distilled by Teachers I$\rightarrow$V$\rightarrow$IV & \bf{37.0} & \bf{56.8} & \bf{39.9}& \bf{20.2}& \bf{39.8} & \bf{50.0} & \bf{33.7} &  \bf{53.6} &\bf{36.0} &\bf{17.2} &\bf{36.0} &\bf{47.3} \\
\bottomrule
\end{tabular}
    }
\label{tab:res18-ablations}
\end{table*}

\noindent{\em The sequential order of the teachers plays a more critical role in the heterogeneous setting.}
Table~\ref{tab:res18-ablations} (rows 7-12) presents representative results with different orders or combinations of the teachers. Again, a proper progressive distillation (row 12) outperforms simultaneous distillation (rows 7-9). Notably, it is necessary to start with Teacher I, since the capacity gap between Student III and Teacher I is minimal, with difference only on the depth of their ResNet backbones. These results confirm the importance of our curriculum-like progression to best benefit from multiple teachers.

\noindent{\bf Training a student longer vs. distilling a student:} As another sanity check, Table~\ref{tab:res18-ablations} includes results of training Student III with more epochs without distillation (rows 2-4). We can see that the first 12 additional epochs improve APs by $1\%$, but there are no significant improvements even if we train for a longer period. This shows the effectiveness of detector distillation.

\noindent{\bf Distillation with different model resolutions:} In Table~\ref{tab:res18-ablations}, we have performed distillation where the student and teacher models operate on the same input image resolution (e.g., the standard resolution $1,333\times 800$ on MS COCO). In practice, one way to further reduce the latency/runtime of the student is to operate on lower-resolution images. However, this poses additional challenges -- with a teacher of high input resolution and a student of low input resolution, they become even more heterogeneous. Moreover, image resolution substantially affects object detection performance~\citep{ashraf2016shallow}. Here, we are interested in performing distillation with models trained with images of different resolutions to further investigate the generalizability of MTPD. More specifically, we use high-resolution models as teachers and low-resolution models as students, as shown in Table~\ref{tab:models-reso} (rows 1-4).

\begin{table*}[t]
    \centering
\caption{Detectors trained with different input resolutions on the COCO dataset. We use a series of Teacher I variants: Teacher I-1 is trained with the standard input resolution of $1,333\times 800$; Teacher I-2 is trained with $1,000\times 600$ input; Teacher I-3 is trained with $666\times 400$ input; and the student is trained with $333\times 200$ input. We report the detection (`Box') and segmentation (`Mask') APs and runtime. We compare our distilled student with its teachers and off-the-shelf (‘OTS’) student. Our approach is {\em effective with even more heterogeneous teacher and student models of different input resolutions}.}
    \resizebox{\columnwidth}{!}{
        \begin{tabular}{ @{}c | l|c|c c c c c c|c c c c c c|c @{} }
            \toprule
           \multirow{2}{*}{ID}& \multirow{2}{*}{Model} & {Input} & \multicolumn{6}{c|}{Box} & \multicolumn{6}{c|}{Mask} & Runtime \\
             & &Resolution & AP & AP$_{50}$ & AP$_{75}$ & AP$_S$ & AP$_M$ & AP$_L$ & AP & AP$_{50}$ & AP$_{75}$ & AP$_S$ & AP$_M$ & AP$_L$ & (ms)\\
              \midrule
    1 & Teacher I-1 & $1333\times 800$ & 38.2 & 58.8 & 41.4& 21.9& 40.9 & 49.5 & 34.7 & 55.7& 37.2& 18.3& 37.4& 47.2& 31.5 \\
    2 & Teacher I-2 & $1000\times 600$ & 37.2 & 57.7 & 40.5& 19.1& 40.9 & 50.4 & 33.6 & 54.3& 35.9& 15.6& 37.0& 47.7& 24.9 \\
    3 & Teacher I-3 & $666\times 400$ & 34.7& 54.0 & 37.2& 15.6& 38.1 & 50.4 & 31.2 & 50.5& 33.2& 12.2& 34.4& 47.0& 19.7 \\
    \midrule
    4 & Student (OTS) & $333\times 200$ & 25.8 & 41.9 & 27.1& 7.0& 27.8 & 44.3 & 23.0 & 38.7& 23.7& 5.0& 23.7& 41.3& 16.9 \\
    5 & Student (Ours) & $333\times 200$ & {\bf 31.5} & {\bf 49.8} & {\bf 33.3}& {\bf 12.3}& {\bf 34.3} & {\bf 48.9} & {\bf 28.2} & {\bf 46.5}& {\bf 29.0} & {\bf 9.3}& {\bf 30.3}& {\bf 45.4}& 16.9 \\
    \bottomrule
\end{tabular}}
\label{tab:models-reso}
\end{table*}

In these experiments, the teacher and student feature maps have different \emph{spatial resolutions}. To tackle this, we simply upsample the spatial maps of the student and supervise the student with the teachers' features. Again, Table~\ref{tab:models-reso} shows that MTPD is effective in this more challenging scenario. Our best performance is achieved by progressively distilling the student with its Teachers I-3, I-2, and I-1.

\section{Generalizability to Other Datasets and Evaluation Protocols}
\label{sec:supp-other-datasets}

In this section, we study the generalizability of MTPD. As an extension from the gold-standard COCO benchmark, we evaluate our distilled student (trained on COCO) on another dataset, Argoverse-HD~\cite{chang2019argoverse}, and with another metric, streaming accuracy~\cite{li2020towards}, and perform distillation on Argoverse-HD directly.

\begin{table*}[htbp]
    \centering
    \caption{Generalizability on Argoverse-HD. On the {\bf left}, we report standard detection accuracy. `OTS' and distilled students are trained on COCO. We observe  $2\%$ AP gains through distillation, even on novel test sets. On the {\bf right}, we report streaming detection accuracy as defined in ~\citet{li2020towards}, in the detection-only setting on a Tesla V100 GPU. The second column denotes the optimal input resolution (that maximizes streaming accuracy). First, we discover that a lighter model and full-resolution input is much more helpful than having an accurate but complex model that needs to downsize input resolution. Second, MTPD further improves over the lightweight model. }
    \resizebox{.495\textwidth}{!}{
    \begin{tabular}{ l| l|l|l|l|l|l|l }
        \toprule
        \multicolumn{2}{c|}{Model} & box AP &  AP$_{50}$ &  AP$_{75}$ &  AP$_S$ &  AP$_M$ & AP$_L$\\
        \midrule
        \multirow{2}{*}{Stud. II} & OTS & 32.7  & 52 & 34.5 & 14.7& 35.8 & 52.8\\
        & Distilled & \bf{34.4} & \bf{54.2} & \bf{35.9}& \bf{15.0}& \bf{36.8}& \bf{57.7}\\ 
        \midrule
       \multirow{2}{*}{Stud. III} & OTS & 28.9 & 48.8 & 30.0 & 12.8& 31.3 & 49.2\\
        & Distilled & \bf{30.6} & \bf{49.7} & \bf{31.8} & \bf{12.9}& \bf{32.6} & \bf{51.9}\\ 
        \bottomrule
    \end{tabular}}
        \resizebox{.495\textwidth}{!}{
\begin{tabular}{@{}lccccccc@{}}
\toprule
Detector                       & Input & AP   & AP$_{50}$ & AP$_{75}$ & AP$_S$ & AP$_M$ & AP$_L$ \\ \midrule
Cas. MRCNN50 \citep{li2020towards}      & 0.5$\times$         & 14.0 & 26.8      & 12.2      & 1.0    & 9.9    & 28.8   \\
MRCNN18 (Ours)    & 1.0$\times$         & 23.7 & 44.8      & 22.6      & 10.4   & 23.1   & 37.8   \\
MRCNN18 (+ Distill) & 1.0$\times$         & {\bf 25.0} & {\bf 45.8}      & {\bf 24.2}      & {\bf 10.5}   & {\bf 24.1}   & {\bf 39.3}  \\ 
\bottomrule
\end{tabular}}
    \label{tab:argoverse}
\end{table*}

\begin{table*}[h!]
    \centering
\caption{Heterogenous distillation of Argoverse-HD detectors, where a student with a ResNet-18 backbone is distilled with teachers with ResNet-50 backbones. We report the detection (`Box') APs and runtime. We compare our distilled student with its teachers and off-the-shelf (`OTS') student. Our distilled student significantly improves the APs over the `OTS' student by over 2\%. Notably, our distilled student achieves detection accuracy that is {\em comparable with Teacher A but with only around third of the runtime}.}
    \resizebox{\columnwidth}{!}{
        \begin{tabular}{ @{}c | l|l l l |c c c c c c|c @{} }
            \toprule
           \multirow{2}{*}{ID}& \multirow{2}{*}{Model} & \multirow{2}{*}{Backbone} & \multirow{2}{*}{Neck} & \multirow{2}{*}{Detection Head} & \multicolumn{6}{c|}{Box} & Runtime \\
             & & & & & AP & AP$_{50}$ & AP$_{75}$ & AP$_S$ & AP$_M$ & AP$_L$  & (ms)\\
              \midrule
    1 & Teacher A & ResNet-50 & FPN & Faster R-CNN & 29.6 &  48.2& 30.5 & 16.4 & 33.1 & 45.1 &  79.2\\
    2 & Teacher B & ResNet-50 & FPN & Cascade & 32.3 & 50.4 & 35.0 & 16.4 & 37.1  & 47.7 & 89.0\\
    3 & Teacher C & ResNet-50 + SAC & RFP & Faster R-CNN & 32.9 & 51.0 & 35.5 & 17.6 & 33.7  & 52.9 & 230.8 \\
    4 & Teacher D & ResNet-50 + SAC & RFP & Cascade & 34.5 & 52.0 & 37.7 & 17.9 &  37.0 & 52.8 & 241.2\\
    \midrule
    5 & Student (OTS) & ResNet-18 & FPN & Faster R-CNN &  27.1 & 48.1 & 27.5& 14.4& 31.2 & 40.0 & 29.3\\
    6 & Student (distilled) & ResNet-18 & FPN & Faster R-CNN & {\bf 29.2} & {\bf 49} & {\bf 30.9} & {\bf 15} & {\bf 31.7} & {\bf 45.6} & 29.5 \\
    \bottomrule
\end{tabular}}
\label{tab:argoverse-distill}
\end{table*}

\noindent{\bf Argoverse-HD} is a more challenging dataset than COCO, due to higher resolution images and significantly more small objects. Constructed from the autonomous driving dataset Argoverse 1.1~\citep{chang2019argoverse}, Argoverse-HD contains RGB video sequences and dense 2D bounding box annotations (1,260k boxes in total). 
It consists of 8 object categories, which are a subset of 80 COCO classes and are directly relevant to autonomous driving: person, bicycle, car, motorcycle, bus, truck, traffic light, and stop sign. There are 38k training images and 15k validation images. We report results on the validation images. We test the distilled models trained on COCO on Argoverse-HD \emph{without re-training}. Table~\ref{tab:argoverse}-left shows the generalizability of MTPD.

\noindent{\bf Streaming accuracy} is a recently proposed metric that simultaneously evaluates both the accuracy and latency of algorithms in an online real-time setting~\citep{li2020towards}. 
The evaluator queries the state of the world at all time instants, forcing algorithms to consider the amount of streaming data that must be ignored while processing the last frame. Following the setup proposed in~\citet{li2020towards}, we evaluate streaming AP in the context of real-time object detection for autonomous vehicles.
Table~\ref{tab:argoverse}-right shows MTPD outperforming the prior results from \citet{li2020towards} by a dramatic margin. 
We find significant wins by using an exceedingly lightweight network (ResNet-18 based Mask R-CNN) that can process full-resolution images without sacrificing latency. Due to much higher quantities of small objects, high-reslution processing is more effective than deeper network structures.  
In addition, progressive distillation further improves performance.

\noindent{\bf Direct distillation on Argoverse-HD:} Given the generalizability of the already-distilled models, now we {\em directly distill} the student model on Argoverse-HD, using Faster R-CNN with a ResNet-18 backbone as the student model. As shown in Table~\ref{tab:argoverse-distill}, we use four teachers with ResNet-50 backbones (rows 1-4), including Faster R-CNN~\citep{ren2015faster} (Teacher A), Cascade R-CNN~\citep{cai2018cascade} (Teacher B), and DetectoRS~\citep{qiao2020detectors} (Teachers C \& D).

The results are summarized in Table~\ref{tab:argoverse-distill}. Our best distillation performance is achieved when we first distill the student with a similar teacher (Teacher A), and then progressively distill with more powerful teachers (Teachers B, then C, and finally D). Overall, the box AP is improved from $27.1\%$ to $29.2\%$.

In addition, comparing with Table~\ref{tab:argoverse}-left, the {\em absolute} performance of the teachers and students in Table~\ref{tab:argoverse-distill} is lower. This is because here we use weaker teachers and student models (Faster R-CNN for fast experiments) than the models used in Table~\ref{tab:argoverse}-left (Mask R-CNN). However, the {\em relative} improvement (between the distilled and OTS students) of box AP ($2.1\%$) is larger than that in Table~\ref{tab:argoverse}-left ($1.7\%$), indicating that learning distillation directly on Argoverse-HD further improves the performance.

\section{Additional Related Work: Detailed Comparison with Prior Knowledge Distillation Methods for Image Classification}
\label{sec:supp-related}
The most profound difference between this work and most of the prior work on knowledge distillation is that prior work mainly focuses on the image classification task, while we address the object detection task. The detection task (and the associated model architectures) is much more complicated than the classification task. This makes the distillation methods developed in the context of classification often not directly applicable to detection. That is why dedicated distillation methods~\citep{chen2017learning,wang2019distilling,zhang2020improve,guo2021distilling,guo2021distillingimage,dai2021general,yang2022focal,yang2022masked} need to be developed for the detection task in the literature. Here, we discuss the difference between our method, Multi-Teacher Progressive Distillation (MTPD), and prior methods on knowledge distillation in detail:

    \noindent{\bf Progressive distillation:} Teacher Assistant Knowledge Distillation (TAKD)~\citep{mirzadeh2020improved} is related to our method, in the sense that this work progressively distills a student from multiple teachers (one teacher and several additional teacher assistants (TAs)). With one TA, the distillation process in TAKD contains three steps: 1) The TA is first distilled from the teacher; 2) The student is distilled from the TA; and 3) The student trained by the TA is further distilled from the teacher. When there are multiple TAs, the shallower TAs are distilled from deeper ones, so that they form a distillation path. However, our work is different from TAKD in three important ways:
    \begin{itemize}[leftmargin=*, noitemsep, nolistsep]
        \item As mentioned above, TAKD focuses on image classification, while we study progressive distillation in the context of object detection. In our case, the transferred knowledge is no longer classification logits but intermediate features or structured predictions. This would require additional consideration and algorithmic designs to extend progressive distillation from image classification to object detection.
        \item Our strategy to construct the teacher sequence is a novel contribution and is fundamentally different from TAKD. In our work, we propose a heuristic algorithm BGS (Algorithm~\ref{alg:teacher-order}) based on the representation similarities between different models (Section~\ref{subsec:multiple}), which automatically generates the teacher order. In TAKD, a series of deep networks with increasing depths act as the student, the TA(s), and the teacher. One can intuitively determine the distillation sequence of TA(s) according to their increasing depths (which imply learning capacities). However in our case, there is a pool of teachers with diverse architectures and their relative ordering is unknown. This challenge motivates us to design an algorithm to automatically decide the teacher order based on their representation similarities; importantly, the strategy in TAKD is not applicable in our task.
        \item TAKD is more cumbersome and time-consuming. An intermediate TA in TAKD needs to be first distilled from the teacher or a deeper TA, so the TAs have to be trained one by one. By contrast, all of our teachers (including the intermediate teachers and the final teacher) are trained independently. This makes the generation of our teachers parallelizable.
    \end{itemize}
    \noindent{\bf Multi-teacher distillation:} The key difference lies in: These methods~\citep{you2017learning,lan2018knowledge,guo2020online} use an ensemble of multiple teachers simultaneously to guide the student learning, while our work distills from multiple teachers sequentially, and we propose a novel method to construct the appropriate teacher order. Empirically, we compare these strategies, and demonstrate that MTPD outperforms the simultaneous strategy (via teacher ensemble by taking the average of teacher features) for object detection.
    \begin{itemize}[leftmargin=*, noitemsep, nolistsep]
        \item This comparison is provided in Section~\ref{sec:supp-ablation-homo} (Table~\ref{tab:res50-combine}) and Section~\ref{sec:supp-ablation-hetero} (Table~\ref{tab:res18-ablations}). For example in Table~\ref{tab:res18-ablations}, if we compare experiments with IDs 7-9 (simultaneous distillation from teacher ensembles) and experiments with IDs 10-12 (progressive distillation from teacher sequences), we find that progressive distillation is a better choice.
        \item Our performance superiority is because in the object detection task, the teacher's knowledge is transferred from intermediate features, rather than from final classification predictions. Thus, the ensemble of multiple teachers might provide conflicting supervision signals for the student, leading to performance interior to our progressive distillation.
    \end{itemize}
    \noindent{\bf Online distillation and deep mutual learning:} Although the teacher model is also changing during online distillation and deep mutual learning~\citep{yang2019snapshot,guo2020online,yao2020knowledge,li2022online}, the principle of our sequential teachers is significantly different from online distillation for the following reasons:
    \begin{itemize}[leftmargin=*, noitemsep, nolistsep]
        \item Strictly speaking, in online distillation, there is only one teacher -- This teacher’s architecture is fixed, and its weights keep updating in an online manner. By contrast, we have multiple teachers -- These teachers have different architectures, and their weights are first trained independently, and then frozen in the progressive distillation process; in our progressive distillation, we switch the whole teacher model.
        \item The type of discrepancy between the student and the teacher is different for ours and online distillation. Online distillation often uses similar or even the same architecture for both the teacher and student models. Consequently, their capacities are at the same level, and they can evolve together. Our study is quite different: The key question we want to address is the capacity gap between the student and the teacher (the capacity gap is due to the architectural difference between the student and the teacher); and our solution is to progressively distill using other teachers with intermediate capacities.
    \end{itemize}

    \noindent{\bf Other distillation mechanisms for classification:} These methods~\citep{romero2014fitnets,zagoruyko2016paying,ahn2019variational} introduce other types of distillation mechanisms, but they are not directly applicable to our setting of multi-teacher detector distillation:
    \begin{itemize}[leftmargin=*, noitemsep, nolistsep]
        \item They still consider the setting where only one single fixed teacher is involved. Different from these methods, we use multiple teachers to progressively transfer knowledge from them to the student.
        \item We do share some similarities with \citet{romero2014fitnets,zagoruyko2016paying,ahn2019variational} in that they are distilling knowledge from the activations of intermediate layers. The simple feature-matching loss (Section~\ref{subsec:single}) and other recent work in detector distillation (e.g., CWD~\citep{shu2021channel}, FGD~\citep{yang2022focal}, and MGD~\citep{yang2022masked}) are based on the ``hint'' distillation (learning from intermediate layers' outputs) from \citet{romero2014fitnets}. However, directly applying such distillation methods designed for classification to object detectors cannot lead to satisfactory student performance, as compared with dedicated detector distillation methods.
        \item We take the best-performing method (among classification-oriented, feature-based distillation methods) Variational Information Distillation (VID)~\citep{ahn2019variational} as an example for comparison with detector distillation methods. The results are shown in Table~\ref{tab:classification-detection}. In this distillation setting with RetinaNet object detectors, VID fails to achieve comparable performance with detector-specific methods such as CWD, FGD, and MGD, and falls even farther behind MGD integrated with MTPD. As discussed previously, we believe that object detection introduces additional complexity to this distillation task, so it requires some customized mechanisms for successful knowledge transfer from the teacher detector to the student. As a result, a distillation loss designed for classification (e.g., VID) is not the optimal solution.
    \end{itemize}

\begin{table*}[t!]
    \centering
    \caption{Learning lightweight RetinaNet detectors with various distillation mechanisms. The classification-oriented distillation mechanism VID fails to achieve comparable performance with detector-specific methods including CWD, FGD, and MGD, and much worse than the combination of MGD and MTPD.}
    \begin{tabular}{@{}c | l | l | l | c@{}}
    \toprule
    ID & Student & Distillation & Teacher(s) & AP \\
    \midrule
    1 & \multirow{1}{*}{RetinaNet/ResNet-50} & VID~\citep{ahn2019variational} & RetinaNet/ResNeXt-101 & 40.3 \\
    \midrule
    2 & \multirow{3}{*}{RetinaNet/ResNet-50} & CWD~\citep{shu2021channel} & RetinaNet/ResNeXt-101 & 40.8 \\
    3 & & FGD~\citep{yang2022focal} & RetinaNet/ResNeXt-101 & 40.7 \\
    4 & & MGD~\citep{yang2022masked} & RetinaNet/ResNeXt-101 & 41.0 \\
    \midrule
    \multirow{2}{*}{5} & \multirow{2}{*}{RetinaNet/ResNet-50} & MGD~\citep{yang2022masked} & RetinaNet/ResNet-101 & \multirow{2}{*}{\bf 41.4} \\
    & & + MTPD (Ours) & $\rightarrow$RetinaNet/ResNeXt-101 & \\
    \bottomrule
    \end{tabular}
    \label{tab:classification-detection}
\end{table*}

\section{Additional Results on Generalizability to State-of-the-Art Distillation Mechanisms}
\label{sec:supp-comb}
In this section, we include additional experimental results to discuss the generalizability of MTPD with a focus on state-of-the-art distillation mechanisms.

\noindent{\bf Transformer-based students:} In the main paper, we have shown that MTPD is the key to successful distillation from Transformer-based teachers to convolution-based students. We prefer {\em not} to distill Transformer-based students from convolution-based teachers for the following reasons:

\begin{itemize}[leftmargin=*, noitemsep, nolistsep]
    \item Consistency with practical scenarios: The end goal of this work and knowledge distillation in general is to develop high-accuracy, lightweight object detectors. To this end, the student detectors are typically efficient detectors like RetinaNet with a ResNet backbone. Although there has been some recent research focusing on designing computationally efficient architectures for Transformers, the widely adopted vision Transformers (e.g., ViT and Swin Transformer) are still slower than their convolution-based counterparts (e.g., ResNet and EfficientNet), because convolution operations are highly optimized on GPUs. Therefore, ResNet-based students are more favored in developing lightweight object detectors and are thus used in our evaluation.

    \item One may consider it more approachable to transfer knowledge from convolution-based teachers to Transformers as CNNs are more matured. However, when developing lightweight object detectors, we usually care less about the training efforts of the teachers, as long as the students are computation-efficient and accurate at test time. The setting where distillation happens from high-capacity Transformer teachers to low-capacity convolution-based students is thus more challenging and more important for real-world applications, than the case of distillation from convolution-based teachers to Transformer-based students.

    \item The key question we want to address is the capacity gap between the student and the teacher. Typically, commonly used Transformer-based backbones (e.g., Swin-Small) have stronger performance than convolution-based backbones (e.g., ResNet-101), and thus possess a higher learning capacity. We study the more challenging case of distilling knowledge from a high-capacity Transformer-based teacher (for better performance) to a low-capacity convolution-based student (for better efficiency), so the capacity gap is unavoidable and must be mitigated. In such a scenario, MTPD is a helpful solution. In fact, we have shown that progressive distillation is critical to the success of knowledge transfer from a Swin Transformer teacher to a ResNet student in Table~\ref{tab:swin}.
\end{itemize}

Nevertheless, to further test the generalizability of MTPD, we investigate distillation from convolution-based teachers into a Transformer-based student. The student is a RetinaNet/Swin-Tiny detector. As for the teachers, we choose detectors with the very recent, strong, convolution-based backbone: ConvNeXt~\citep{liu2022convnet}. We use the state-of-the-art distillation loss MGD as in the main paper. The comparison of the results are shown in Table~\ref{tab:convnext}. In this convolution-to-Transformer distillation setting, MGD direct distillation using a strong ConvNeXt-Small backbone gives the Transformer-based RetinaNet/Swin-Tiny student a 0.9\% AP improvement. MTPD via an intermediate ConvNeXt-Tiny teacher further boosts the student performance by 0.6\% AP and leads to a high-accuracy student with 45.2\% AP on COCO detection. This experiment again demonstrates the effectiveness of MTPD across various teacher/student architectures, in addition to the Transformer-teacher, convolution-student setting which we have shown in the main paper.

\begin{table*}[t!]
    \centering
    \caption{Distillation with a Transformer-based student. In contrast to the Transformer-teacher, convolution-student setting in our main paper, here ConvNeXt-backboned~\cite{liu2022convnet} teachers are employed to teach a Swin-backboned~\cite{liu2021swin} student. Again, MTPD (ID 3) leads to the best student accuracy.}
    \resizebox{\columnwidth}{!}{
    \begin{tabular}{@{}c | l | l | l | c | c@{}}
    \toprule
    ID & Student & Distillation & Teacher(s) & Training Schedule &AP \\
    \midrule
    1 & \multirow{2}{*}{RetinaNet/Swin-Tiny} & None & None & $2\times$ & 43.7 \\
    2 & & MGD~\citep{yang2022masked} & Cascade Mask R-CNN/ConvNeXt-Small & $2\times$ & 44.6 \\
    \midrule
    \multirow{2}{*}{3} & \multirow{2}{*}{RetinaNet/Swin-Tiny} & MGD~\citep{yang2022masked} & Mask R-CNN/ConvNeXt-Tiny & \multirow{2}{*}{$1\times$ $+$ $1\times$} & \multirow{2}{*}{\bf 45.2} \\
    & & + MTPD (Ours) & $\rightarrow$Cascade Mask R-CNN/ConvNeXt-Small & & \\
    \bottomrule
    \end{tabular}}
    \label{tab:convnext}
\end{table*}

\noindent{\bf Keeping using the intermediate teacher:} In the main paper, we have discussed the capacity gap between the teacher and student and how to mitigate this gap via progressive distillation. One may question that, since the capacity gap between the \emph{intermediate teacher} and the student is smaller, keeping using this intermediate teacher throughout the distillation procedure might also lead to a good student. Here we provide an exemplary experiment result (in addition to Figure~\ref{fig:comb}a) as an answer to this question: Only using the intermediate teacher is still suboptimal as compared with MTPD.

\begin{table*}[t!]
    \centering
    \caption{Distillation of RetinaNet detectors. The capacity gap between the student (RetinaNet/ResNet-50) and the intermediate teacher (RetinaNet/ResNet-101) is smaller. MTPD is better than both distillation schemes that keep using the intermediate teacher (ID 3) or the final teacher (ID 2).}
    \resizebox{\columnwidth}{!}{
    \begin{tabular}{@{}c | l | l | l | c | c@{}}
    \toprule
    ID & Student & Distillation & Teacher(s) & Training Schedule &AP \\
    \midrule
    1 & RetinaNet/ResNet-50 & None & None & $2\times$ & 37.4 \\
    \midrule
    2 & \multirow{3}{*}{RetinaNet/ResNet-50} & \multirow{3}{*}{MGD~\citep{yang2022masked}} & RetinaNet/ResNeXt-101 & $2\times$ & 41.0 \\
    3 & & & RetinaNet/ResNet-101 & $2\times$ & 40.7 \\
    4 & & & RetinaNet/ResNet-101 & $1\times$ & 40.2 \\
    \midrule
    \multirow{2}{*}{5} & \multirow{2}{*}{RetinaNet/ResNet-50} & MGD~\citep{yang2022masked} & RetinaNet/ResNet-101 & \multirow{2}{*}{$1\times$ $+$ $1\times$} & \multirow{2}{*}{\bf 41.4} \\
    & & + MTPD (Ours) & $\rightarrow$RetinaNet/ResNeXt-101 & & \\
    \bottomrule
    \end{tabular}}
    \label{tab:intermidiate}
\end{table*}

In this experiment, we use state-of-the-art distillation method MGD as the base method. We use RetinaNet/ResNet-50 (37.4 AP on COCO) as the student model, RetinaNet/ResNet-101 (38.9 AP) as the intermediate teacher model, and RetinaNet/ResNeXt-101 (40.8 AP) as the final teacher model. The capacity gap between the intermediate teacher and the student is smaller than that between the final teacher and the student. Table~\ref{tab:intermidiate} shows the results.

Keeping using the intermediate teacher (RetinaNet/ResNet-101) for a longer $2\times$ training schedule indeed improves the performance from 40.2\% AP to 40.7\%, but it is still not better than directly using the final teacher (RetinaNet/ResNeXt-101). Our progressive distillation, which first uses the intermediate teacher and then the final teacher for distillation, outperforms both direct distillation schemes and achieves \textbf{41.4\%} AP performance. As always, we use the same total training time ($2\times$ training schedule) as direct distillation for a fair comparison. This experiment supports the conclusions that 1) employing an intermediate teacher throughout the distillation process is not a good option; and 2) MTPD, which uses both the intermediate teacher and the best performing teacher sequentially, leads to the best student performance.

\section{Implementation Details}
\label{sec:supp-impl}

We implement detectors and their distillation using the MMDetection codebase~\citep{mmdetection}. We train on $8$ GPUs for $12$ epochs for each distillation.
For MS COCO, we use the standard input resolution of $1,333 \times 800$, with each GPU hosting 2 images.
For Argoverse-HD, we use its much higher native resolution as the input at $1,920 \times 1,200$, with each GPU hosting 1 image.
We use an initial learning rate of $0.01$ (for RetinaNet students) or $0.02$ (for Mask R-CNN students). We use stochastic gradient descent and a momentum of $0.9$.
For the simple feature-matching loss (see Section~\ref{subsec:single}), we perform a grid search over the hyper-parameter $\lambda$. While the optimal values are dependent on the architectures of the teacher and student models, we find that the performance is not very sensitive to $\lambda$ between $0.3$ and $0.8$. We set $\lambda=0.5$ for RetinaNet students and $\lambda=0.8$ for Mask R-CNN students.

When we integrate MTPD with state-of-the-art distillation mechanisms including CWD~\citep{shu2021channel}, FGD~\citep{yang2022focal}, and MGD~\citep{yang2022masked} (Section~\ref{subsec:combination}), we strictly follow the publicly available implementation from their authors, and use an intermediate teacher (RetinaNet/ResNet-101 or Cascade Mask R-CNN/ResNet50-DCN) for progressive distillation. In the original implementation of FGD and MGD, an inheriting strategy~\citep{kang2021instance} is utilized, which initializes the student with the teacher’s neck and head parameters to train the student when they have the same head structure. In MTPD, we adopt this inheriting strategy only once for the first teacher.

For the Transformer-based teachers, we use the Swin Transformer backbone, which has a hierarchical architecture and shares the ``same feature map resolutions as those of typical convolutional networks (e.g., ResNet-50)''~\citep{liu2021swin}. Following the original implementation of Swin Transformer, the backbone is equipped with an FPN neck, so the number of neck feature channels is the same as the student. As a result, Swin Transformer based teachers can be used like some other convolution-based teachers without the feature map matching function ($r(\cdot)$ in Section~\ref{subsec:single}).


\end{document}